\documentclass{article} 
\usepackage{iclr2021_conference,times}

\usepackage{amsmath,amsfonts,bm}

\def\1{\bm{1}}

\DeclareMathAlphabet{\mathsfit}{\encodingdefault}{\sfdefault}{m}{sl}
\SetMathAlphabet{\mathsfit}{bold}{\encodingdefault}{\sfdefault}{bx}{n}

\DeclareMathOperator*{\argmin}{arg\,min}

\usepackage{hyperref}
\usepackage{url}

\usepackage{microtype}
\usepackage{graphicx}
\usepackage{subfigure}
\usepackage{booktabs} 
\usepackage{bm}
\usepackage{stmaryrd}

\usepackage{caption}
\usepackage{graphicx}
\usepackage{enumitem}
\usepackage{algorithm}
\usepackage{algorithmic}
\usepackage{wrapfig}

\usepackage{hyperref}

\newcommand{\lgbm}{\textbf{LightGBM}}
\newcommand{\gbdt}{\textbf{CatBoost}}
\newcommand{\mlp}{\textbf{FCNN}}
\newcommand{\gnn}{\textbf{GNN}}
\newcommand{\bgnn}{\textbf{BGNN}}
\newcommand{\resgnn}{\textbf{Res-GNN}}
\newcommand{\gat}{\textbf{GAT}}
\newcommand{\gcn}{\textbf{GCN}}
\newcommand{\agnn}{\textbf{AGNN}}
\newcommand{\appnp}{\textbf{APPNP}}
\newcommand{\dt}{\textbf{GBDT}}

\newcommand{\house}{\textbf{House}}
\newcommand{\county}{\textbf{County}}
\newcommand{\vkdata}{\textbf{VK}}
\newcommand{\wiki}{\textbf{Wiki}}
\newcommand{\avazu}{\textbf{Avazu}}
\newcommand{\arxiv}{\textbf{OGB-ArXiv}}
\newcommand{\slap}{\textbf{Slap}}
\newcommand{\dblp}{\textbf{DBLP}}

\usepackage[section]{placeins}
\usepackage{array,multirow,graphicx}
\newcommand{\STAB}[1]{\begin{tabular}{@{}c@{}}#1\end{tabular}}

\usepackage{amsmath}
\usepackage{setspace}
\usepackage{mathrsfs}
\usepackage{amsfonts}

\newcommand{\x}{\mathbf{x}}
\newcommand{\X}{\mathbf{X}}

\title{Boost then Convolve: Gradient Boosting \\Meets Graph Neural Networks}

\author{Sergei Ivanov \\
Criteo AI Lab\\
Paris, France \\
\texttt{s.ivanov@criteo.com} \\
\And
Liudmila Prokhorenkova \\
Yandex; HSE University; MIPT \\
Moscow, Russia \\
\texttt{ostroumova-la@yandex-team.ru} \\
}

\iclrfinalcopy 
\begin{document}

\maketitle

\begin{abstract}
Graph neural networks (GNNs) are powerful models that have been successful in various graph representation learning tasks. Whereas gradient boosted decision trees (GBDT) often outperform other machine learning methods when faced with heterogeneous tabular data. But what approach should be used for graphs with tabular node features? Previous GNN models have mostly focused on networks with homogeneous sparse features and, as we show, are suboptimal in the heterogeneous setting. In this work, we propose a novel architecture that trains GBDT and GNN jointly to get the best of both worlds: the GBDT model deals with heterogeneous features, while GNN accounts for the graph structure. Our model benefits from end-to-end optimization by allowing new trees to fit the gradient updates of GNN. With an extensive experimental comparison to the leading GBDT and GNN models, we demonstrate a significant increase in performance on a variety of graphs with tabular features. The code is available: \url{https://github.com/nd7141/bgnn}.
\end{abstract}

\section{Introduction}

Graph neural networks (GNNs) have shown great success in learning on graph-structured data with various applications in molecular design \citep{stokes2020deep}, computer vision \citep{casas2019spatially}, combinatorial optimization \citep{mazyavkina2020reinforcement}, and recommender systems \citep{sun20framework}. The main driving force for progress is the existence of canonical GNN architecture that efficiently encodes the original input data into expressive representations, thereby achieving high-quality results on new datasets and tasks.  

Recent research has mostly focused on GNNs with sparse data representing either homogeneous node embeddings (e.g., one-hot encoded graph statistics) or bag-of-words representations. Yet tabular data with detailed information and rich semantics among nodes in the graph are more natural for many situations and abundant in real-world AI \citep{xiao2019non}. For example, in a social network, each person has socio-demographic characteristics (e.g., age, gender, date of graduation), which largely vary in data type, scale, and missing values. GNNs for graphs with tabular data remain unexplored, with gradient boosted decision trees (GBDTs) largely dominating in applications with such heterogeneous data~\citep{bentejac2020comparative}. 

GBDTs are so successful for tabular data because they possess certain properties: (i) they efficiently learn decision space with hyperplane-like boundaries that are common in tabular data; (ii) they are well-suited for working with variables of high cardinality, features with missing values, and of different scale; (iii) they provide qualitative interpretation for decision trees (e.g., by computing decrease in node impurity for every feature) or for ensembles via post-hoc analysis stage~\citep{kaur2020interpreting}; (iv) in practical applications, they mostly converge faster even for large amounts of data. 

In contrast, a crucial feature of GNNs is that they take into account both the neighborhood information of the nodes and the node features to make a prediction, unlike GBDTs that require additional preprocessing analysis to provide the algorithm with graph summary (e.g., through unsupervised graph embeddings~\citep{hu2020ogb}). Moreover, it has been shown theoretically that message-passing GNNs can compute any function on its graph input that is computable by a Turing machine, i.e., GNN is known to be the only learning architecture that possesses universality properties  on graphs (approximation~\citep{keriven2019universal, maron2019universality} and computability~\citep{loukas2020what}). Furthermore, gradient-based learning of neural networks can have numerous advantages over the tree-based approach: (i) relational inductive bias imposed in GNNs alleviates the need to manually engineer features that capture the topology of the network \citep{battaglia2018relational}; (ii) the end-to-end nature of training neural networks allows multi-stage~\citep{fey2020deep} or multi-component \citep{wang2020abstract} integration of GNNs in application-dependent solutions; (iii) pretraining representations with graph networks enriches transfer learning for many valuable tasks such as unsupervised domain adaptation \citep{wu20unsupervised}, self-supervised learning \citep{hu2020strategies}, and active learning regimes~\citep{garcia2018fewshot}. 

Undoubtedly, there are major benefits in both GBDT and GNN methods. Is it possible to get advantages of both worlds? All previous approaches~\citep{arik2020tabnet, popov2019neural, badirli2020gradient} that attempt to combine gradient boosting and neural networks are computationally heavy,  do not consider graph-structured data, and suffer from the lack of relational bias imposed in GNN architectures, see Appendix~\ref{app:related} for a more detailed comparison with related literature. To the best of our knowledge, the current work is the first to explore using GBDT models for graph-structured data. 

In this paper, we propose a novel learning architecture for graphs with tabular data, \bgnn{}, that combines GBDT’s learning on tabular node features with GNN that refines the predictions utilizing the graph’s topology. This allows \bgnn{} to inherit the advantages of gradient boosting methods (heterogeneous learning and interpretability) and graph networks (representation learning and end-to-end training). Overall, our contributions are the following:

\begin{enumerate}[label=(\arabic*)]
    \item We design a novel generic architecture that combines GBDT and GNN into a unique pipeline. To the best of our knowledge, this is the first work that systematically studies the application of GBDT to graph-structured data. 
    \item We overcome the challenge of end-to-end training of GBDT by iteratively adding new trees that fit the gradient updates of GNN. This allows us to backpropagate the error signal from the topology of the network to GBDT. 
    \item We perform an extensive evaluation of our approach against strong baselines in node prediction tasks. Our results consistently demonstrate significant performance improvements on heterogeneous node regression and node classification tasks over a variety of real-world graphs with tabular data. 
    \item We show that our approach is also more efficient than the state-of-the-art GNN models due to much faster loss convergence during training. Furthermore, learned representations exhibit discernible structure in the latent space, which further demonstrates the expressivity of our approach. 
\end{enumerate}

\section{Background}

Let $G = (V, E)$ be a graph with nodes having features and target labels. 
In node prediction tasks (classification or regression), some target labels are known, and the goal is to predict the remaining ones. Throughout the text, by lowercase variables $\x_v$ ($v \in V$) or $\x$ we denote features of individual nodes, and $\X$ represents the matrix of all features for $v \in V$. Individual target labels are denoted by $y_v$, while 
$Y$ is the vector of known labels.

\textbf{Graph Neural Networks (GNNs)} use both the network’s connectivity and the node features to learn latent representations for all nodes $v \in V$. Many popular GNNs use a neighborhood aggregation approach, also called the message-passing mechanism, where the representation of a node $v$ is updated by applying a non-linear aggregation function of $v$’s neighbors representation \citep{fey2019fast}. Formally, GNN is a differentiable, permutation-invariant function $g_{\theta}: (G, \X) \mapsto \widehat{Y}$, where $\widehat{Y}$ is the vector of predicted labels. Similar to traditional neural networks, GNNs are composed of multiple layers, each representing a non-linear message-passing function:
\begin{equation}
\label{eq:gnn}
    \x^{t}_v = \text{COMBINE}^{t}\left(\x^{t-1}_v, 
    \text{AGGREGATE}^{t}\left(\left\{(\x^{t-1}_w, \x^{t-1}_v):  (w, v) \in E\right\}\right)\right),
\end{equation}
where $\x_v^t$ is the representation of node $v$ at layer $t$, and $\text{COMBINE}^t$ and $\text{AGGREGATE}^t$ are (parametric) functions that aggregate representations from the local neighborhood of a node. Then, the GNN mapping $g_{\theta}$ includes multiple layers of aggregation~\eqref{eq:gnn}. Parameters of GNN model $\theta$ are optimized with gradient descent by minimizing an empirical loss function $L_{\mathrm{GNN}}(Y, g_{\theta}(G, \X))$. 


\textbf{Gradient Boosted Decision Trees (GBDT)} is a well-known and widely used algorithm that is defined on non-graph tabular data~\citep{friedman2001greedy} and is particularly successful for tasks containing heterogeneous features and noisy data. 

The core idea of gradient boosting is to construct a strong model by iteratively adding weak ones (usually decision trees). 
Formally, at each iteration $t$ of the gradient boosting algorithm, the model $f(\x)$ is updated in an additive manner:
\begin{equation}
\label{eq:additive}
f^{t}(\x) = f^{t-1}(\x) + \epsilon\, h^{t}(\x),
\end{equation}
where $f^{t-1}$ is a model constructed at the previous iteration, $h^{t}$ is a weak learner that is chosen from some family of functions $\mathcal{H}$, and $\epsilon$ is a learning rate. The weak learner $h^{t}\in \mathcal{H}$ is chosen to approximate the negative gradient of a loss function $L$ w.r.t.~the current model's predictions:
\begin{equation}
\label{eq:tree}
h^{t} = \argmin_{h\in \mathcal{H}} \sum_i \left(-\frac{\partial L(f^{t-1}(\x_i), y_i)}{\partial f^{t-1}(\x_i)}- h(\x_i)\right)^2 . 
\end{equation}
The gradient w.r.t.~the current predictions indicates how one should change these predictions to improve the loss function. Informally, gradient boosting can be thought of as performing gradient descent in functional space. 

The set of weak learners $\mathcal{H}$ is usually formed by shallow decision trees. Decision trees are built by a recursive partition of the feature space into disjoint regions called leaves. This partition is usually constructed greedily to minimize the loss function~\eqref{eq:tree}. Each leaf $R_j$ of the tree is assigned to a value $a_j$, which estimates the response $y$ in the corresponding region. In our case, $a_j$ is equal to the average negative gradient value in the leaf $R_j$. To sum up, we can write $h(x) = \sum_j a_j \mathrm{1}_{\{x \in R_j\}}$. 

\section{GBDT meets GNN}
\label{sec:bgnn}

\begin{tabular}{cc}
\noindent\begin{minipage}{.43\textwidth}
\centering
\includegraphics[width=0.8\textwidth]{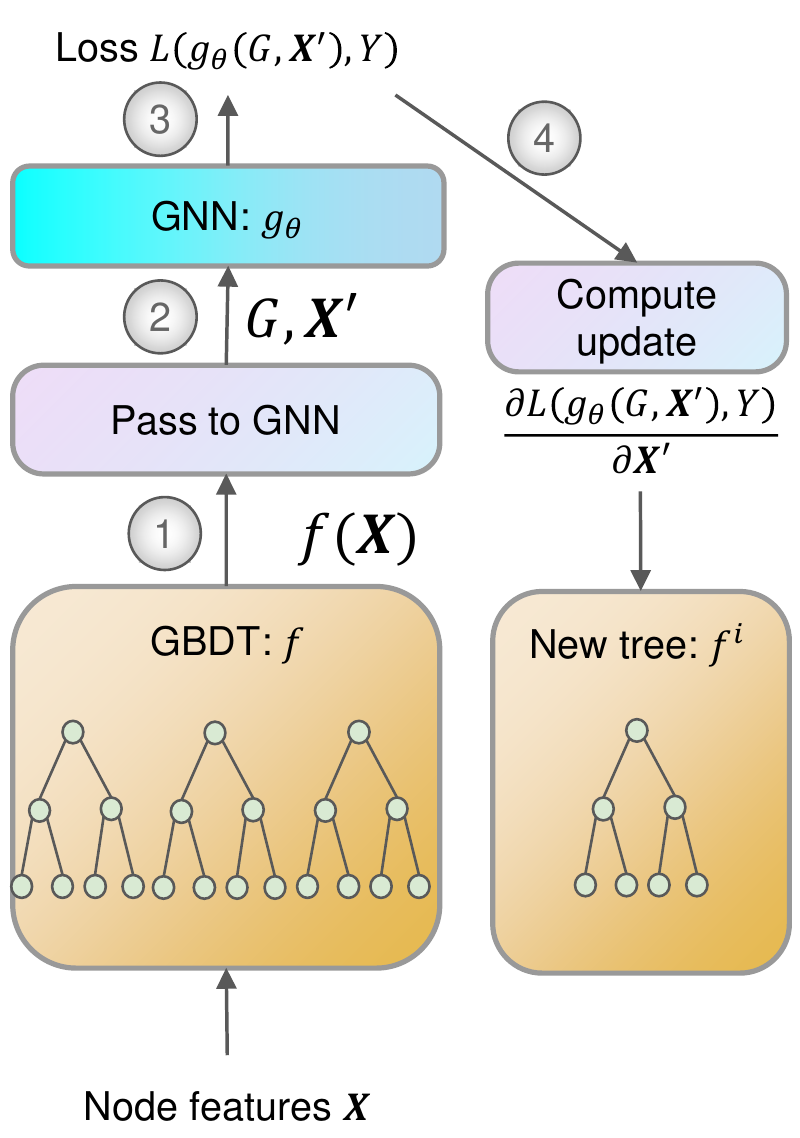}
\captionof{figure}{\footnotesize{Training of \bgnn{}, steps for one epoch are numbered.}}
\label{fig:architecture}
\end{minipage} &
\begin{minipage}{.5\textwidth}
\begin{algorithm}[H]
   \caption{Training of \bgnn{}}
   \label{alg:training}
\begin{algorithmic} 
   \STATE {\bfseries Input:} Graph $G$, node features $\X$, targets $Y$
   
  \STATE Initialize GBDT targets $\mathcal{Y} = Y$
   \FOR{epoch $i=1$ to $N$}
   
   \STATE \# Train $k$ trees of GBDT with eq. \eqref{eq:additive}-\eqref{eq:tree}
   \STATE $f^i \xleftarrow[k]{} \argmin\limits_{f^{i}}L_{\mathrm{GBDT}}(f^{i}(\X), \mathcal{Y})$  \vskip 1pt
   \STATE $f \leftarrow f +  f^{i}$
   
   \STATE
   
   \STATE \# Train $l$ steps of GNN on new node features
   \STATE  $\X' \leftarrow f(\X)$
   \STATE $\theta, \X' \xleftarrow[l]{} \argmin\limits_{\theta, \X'}L_{\mathrm{GNN}}(g_{\theta}(G, \X'), Y)$ \vskip 5pt
   
   \STATE \# Update targets for next iteration of GBDT
   \STATE $\mathcal{Y} \leftarrow \X' - f(\X)$
   
   \ENDFOR
   \STATE {\bfseries Output:} Models GBDT $f$ and GNN $g_{\theta}$
\end{algorithmic}
\end{algorithm}
\end{minipage}
\end{tabular}

Gradient boosting approach is successful for learning on tabular data; however, there are challenges of applying GBDT on graph-structured data: (i) how to propagate relational signal, in addition to node features, to otherwise inherently tabular model; and (ii) how to train it together with GNN in an end-to-end fashion. Indeed, optimizations of GBDT and GNN follow different approaches: the parameters of GNN are optimized via gradient descent, while GBDT is constructed iteratively, and the decision trees remain fixed after being built (decision trees are based on hard splits of the feature space, which makes them non-differentiable).

A straightforward approach would be to train the GBDT model only on the node features and then use the obtained predictions, jointly with the original input, as new node features for GNN. In this case, the graph-insensitive predictions of GBDT will further be refined by a graph neural network. This approach (which we call \resgnn{}) can already boost the performance of GNN for some tasks. However, in this case, the GBDT model completely ignores the graph structure and may miss descriptive features of the graph, providing inaccurate input data to GNN.

In contrast, we propose end-to-end training of GBDT and GNN called \bgnn{} (for Boost-GNN). As before, we first apply GBDT and then GNN, but now we optimize both of them, taking into account the quality of final predictions. The training of \bgnn{} is shown in Figure~\ref{fig:architecture}. Recall that one cannot tune already built decision trees due to their discrete structure, so we iteratively update the GBDT model by adding new trees that approximate the GNN loss function.

In Algorithm~\ref{alg:training}, we present the training of \bgnn{} that combines GBDT and GNN for any node-level prediction problem such as semi-supervised node regression or classification. 
In the first iteration, we build a GBDT model $f^1(\x)$ with $k$ decision trees by minimizing the loss function $L_{\mathrm{GBDT}}(f^1(\x), y)$ (e.g., RMSE for regression or cross-entropy for classification) averaged over the train nodes, following the equations~\eqref{eq:additive}-\eqref{eq:tree}. Using all predictions $f^1(\X)$, we update the node features to $\X'$ that we pass to GNN. Possible update functions that we experiment with include concatenation with the original node features and their replacement by~$f^1(\X)$. 
Next, we train a graph neural network $g_{\theta}$ on a graph $G$ with node features $\X'$ by minimizing $L_{\mathrm{GNN}}(g_{\theta}(G,\X'), Y)$ with $l$ steps of gradient descent.\footnote{$L_{\mathrm{GNN}}$ is determined by the final task, e.g., RMSE for regression or the cross-entropy loss for classification.} Importantly, we optimize both the parameters $\theta$ of GNN and the node features $\X'$. Then, we use the difference between the optimized node features $\X'_{new}$ and the input node features $\X' = f^1(\X)$ as the target for the next decision trees built by GBDT. If $l=1$, the difference $\X'_{new} - \X'$ exactly equals the negative gradient of the loss function w.r.t.~the input features $\X'$ multiplied by the learning rate $\eta$:
\begin{equation*}
    \X'_{new} = \X' - \eta \frac{\partial L_{\mathrm{GNN}}(g_{\theta}(G,\X'), Y)}{\partial \X'}\,.
\end{equation*}
In the second iteration, we train a new GBDT model $f^2$ with the original input features $\X$ but new target labels: $\X'_{new} - \X'$. Intuitively, $f^2$ fits the direction that would improve GNN prediction based on the first predictions $f^1(\X)$. 
In other words, GBDT approximates the gradient steps made by GNN for the node features $\X'$. This is a regression problem, so here $L_{\mathrm{GBDT}}$ is the RMSE loss.

After $f^2$ is trained, we combine the predictions $f(\X) = f^1(\X) + f^2(\X)$ and pass the obtained values $\X'$ to GNN as node features. GNN model $g_{\theta}$ again does $l$ steps of backpropagation and passes the new difference $\X'_{new} - \X'$ as a target to the next iteration of GBDT. In total, the model is trained for $N$ epochs and outputs a GBDT model $f: \X \mapsto Y$ and GNN model $g_{\theta}: (G, \X) \mapsto Y$, which can be used for downstream tasks. 

Intuitively, \bgnn{} model consists of two consecutive blocks, GBDT and GNN, which are trained end-to-end, and therefore can be interpreted from two angles: GBDT as an embedding layer for GNN or GNN as a parametric loss function for GBDT. In the former case, GBDT transforms the original input features $\X$ to new node features $\X'$, which are then passed to GNN. In the latter case, one can see \bgnn{} as a standard gradient boosted training where GNN acts as a complex loss function that depends on the graph topology.

\section{Experiments}

We have performed a comparative evaluation of \bgnn{} and \resgnn{} against a wide variety of strong baselines and previous approaches on heterogeneous node prediction problems, achieving significant improvement in performance across all of them. This section outlines our experimental setting, the results on node regression and classification problems, and extracted feature representations.

In our first experiments, we want to answer two questions:
\begin{enumerate}[label=Q\arabic*]
\item \label{item:qreg1} \textit{Does combination of GBDT and GNN lead to better qualitative results in heterogeneous node regression and classification problems?}
\item \label{item:qreg2} \textit{Is the end-to-end training proposed in Algorithm~\ref{alg:training} better than a combination of pretrained GBDT with GNN?}
\end{enumerate}

To answer these questions, we consider several strong baselines among GBDTs, GNNs, and pure neural networks. \gbdt{} is a recent GBDT implementation \citep{prokhorenkova2018catboost} that uses oblivious trees as weak learners. \lgbm{} is another GBDT model \citep{lightgbm} that is used extensively in ML competitions. Among GNNs, we tested four state-of-the-art recent models that showed superior performance in node prediction tasks: \gat{} \citep{velickovic2018graph}, \gcn{} \citep{kipf2016semi}, \agnn{} \citep{thekumparampil2018attention}, \appnp{} \citep{klicpera2018predict}. Additionally, we test the performance of fully-connected neural network \mlp{} and its end-to-end combination with GNNs, \mlp-\gnn. 

We compare these baselines against two proposed approaches: the end-to-end \bgnn{} model and not end-to-end \resgnn{}. The \bgnn{} model follows Algorithm~\ref{alg:training} and builds each tree approximating the \gnn{} error in the previous iteration. In contrast, \resgnn{} first trains a GBDT model on the training set of nodes and then either appends its predictions for all nodes to the original node features or replaces the original features with the GBDT predictions, after which GNN is trained on the updated features, and GNN's predictions are used to calculate metrics. Hence, \resgnn{} is a two-stage approach where the training of GBDT is independent of GNN. On the other hand, \bgnn{} trains GBDT and GNN simultaneously in an end-to-end fashion. In most of our experiments, the GNN-component of \mlp-\gnn, \resgnn{}, and \bgnn{} is based on \gat{}, while in Section~\ref{sec:expablation} we analyze consistency of improvements across different GNN models.

We ensure that the comparison is done fairly by training each model until the convergence with a reasonable set of hyperparameters evaluated on the validation set. We run each hyperparameter setting three times and take the average of the results. Furthermore, we have five random splits of the data, and the final number represents the average performance of the model for all five random seeds. More details about hyperparameters can be found in Appendix~\ref{app:hyper}.

\subsection{Node regression}

\subsubsection{Datasets}
We utilize five real-world node regression datasets with different properties outlined in Table~\ref{tab:datareg}. Four of these datasets are heterogeneous, i.e., the input features are of different types, scales, and meaning. For example, for the \vkdata{} dataset, the node features are both numerical (e.g., last time seen on the platform) and categorical (e.g., country of living and university). On the other hand, \wiki{} dataset is homogeneous, i.e., the node features are interdependent and correspond to the bag-of-words representations of Wikipedia articles. Additional details about the datasets can be found in Appendix~\ref{app:datasets}.

\begin{table*}[h]
\caption{Summary of regression datasets.}\label{tab:datareg}
\begin{center}
\centering
\footnotesize

\begin{tabular}{lrrrrr}
\toprule
 & \textbf{House} & \textbf{County} & \textbf{VK} & \textbf{Avazu} & \textbf{Wiki} \\
 \midrule
 \midrule
\textbf{Setting} & Heterogeneous & Heterogeneous & Heterogeneous & Heterogeneous & Homogeneous        \\
\textbf{\# Nodes}         &  20640 &	3217 &	54028  &	1297   &  5201       \\
\textbf{\# Edges}         &  182146 &	12684 &	213644 &	54364  &  198493     \\ 
\textbf{\# Features/Node} &  6 &	7 &	14	           &	9      &  3148       \\ 
\textbf{Mean Target}      &   2.06 &	5.44 &	35.47  &	0.08   &  27923.86    \\ 
\textbf{Min Target}       &    0.14 &	1.7	 &  13.48  &	0      &  16          \\
\textbf{Max Target}       &    5.00 &	24.1 &	118.39 &    1      &  849131          \\
\textbf{Median Target}    & 1.79 &	5	 &  33.83      &	0      &  9225        \\
\bottomrule
\end{tabular}
\end{center}
\end{table*}

\subsubsection{Results}
The results of our comparative evaluation for node regression are summarized in Table~\ref{tab:regbalanced}. We report the mean RMSE (with standard deviation) on the test set and the relative gap between RMSE of the \gat{} model \citep{velickovic2018graph} and other methods, i.e., $\mathrm{gap} = (r_m-r_{gnn})/r_{gnn}$, where $r_m$ and $r_{gnn}$ are RMSE of that model and of \gat{}, respectively.

\begin{table*}[h]
\caption{Summary of our results for node regression. Gap \% is relative difference w.r.t.~\gat{} RMSE (the smaller the better). Top-2 results are highlighted in \textbf{bold}.}\label{tab:regbalanced}
\vskip 0.15in
\begin{center}
\centering
\footnotesize
\resizebox{\textwidth}{!}{
\begin{tabular}{ll|rr|rr|rr|rr|rr}
 & & \multicolumn{8}{c|}{\textbf{\textit{Heterogeneous}}} & \multicolumn{2}{c}{\textbf{\textit{Homogeneous}}} \\ \midrule
 & &  \multicolumn{2}{c|}{\textbf{House}} & \multicolumn{2}{c|}{\textbf{County}} & \multicolumn{2}{c|}{\textbf{VK}} & \multicolumn{2}{c|}{\textbf{Avazu}} & \multicolumn{2}{c}{\textbf{Wiki}} \\
 &  Method &  
 \multicolumn{1}{c}{RMSE} & \multicolumn{1}{c|}{Gap \%}  & 
 \multicolumn{1}{c}{RMSE} & \multicolumn{1}{c|}{Gap \%} &
 \multicolumn{1}{c}{RMSE} & \multicolumn{1}{c|}{Gap \%} &
 \multicolumn{1}{c}{RMSE} & \multicolumn{1}{c|}{Gap \%} &
 \multicolumn{1}{c}{RMSE} & \multicolumn{1}{c}{Gap \%} \\
 \midrule
 \midrule
\multirow{2}{*}{\STAB{\rotatebox[origin=c]{90}{\scriptsize{GBDT}}}}
& \gbdt{}      & 0.63 $\pm$ 0.01 & 15.3 & 1.39 $\pm$ 0.07 & -4.32 & 7.16 $\pm$ 0.20 & -0.82 & 0.1172 $\pm$ 0.02 & 3.36 & 46359 $\pm$ 4508 & 0.97    \\
& \lgbm{}        & 0.63 $\pm$ 0.01 & 15.98 & 1.4 $\pm$ 0.07 & -3.93 & 7.2 $\pm$ 0.21 & -0.33 & 0.1171 $\pm$ 0.02 & 3.27 & 49915 $\pm$ 3643 & 8.71     \\
\midrule
\multirow{4}{*}{\STAB{\rotatebox[origin=c]{90}{GNN}}}
& \gat{}          & 0.54 $\pm$ 0.01 & 0 & 1.45 $\pm$ 0.06 & 0 & 7.22 $\pm$ 0.19 & 0  & 0.1134 $\pm$ 0.01 & 0 & \textbf{45916 $\pm$ 4527} & \textbf{0}  \\
& \gcn{}          & 0.63 $\pm$ 0.01 & 16.77 & 1.48 $\pm$ 0.08 & 2.06 & 7.25 $\pm$ 0.19 & 0.34 & 0.1141 $\pm$ 0.02 & 0.58 & \textbf{44936 $\pm$ 4083} & \textbf{-2.14}   \\
& \agnn{}         & 0.59 $\pm$ 0.01 & 8.01 & 1.45 $\pm$ 0.08 & -0.19 & 7.26 $\pm$ 0.20  & 0.54  & 0.1134 $\pm$ 0.02 & -0.02 & 45982 $\pm$ 3058 & 0.14      \\
& \appnp{}        & 0.69 $\pm$ 0.01 & 27.11 & 1.5 $\pm$ 0.11 & 3.39 & 13.23 $\pm$ 0.12 & 83.19 & 0.1127 $\pm$ 0.01 & -0.65 & 53426 $\pm$ 4159 & 16.36   \\
\midrule
\multirow{2}{*}{\STAB{\rotatebox[origin=c]{90}{NN}}}
& \mlp{}            & 0.68 $\pm$ 0.02 & 25.49 & 1.48 $\pm$ 0.07 & 1.56 & 7.29 $\pm$ 0.21 & 1.02 & 0.118 $\pm$ 0.02 & 4.07   & 51662 $\pm$ 2983 & 12.51 \\
& \mlp-\gnn         & 0.53 $\pm$ 0.01 & -2.48 & 1.39 $\pm$ 0.06 & -4.68 & 7.22 $\pm$ 0.20 & 0.01 & 0.1114 $\pm$ 0.02 & -1.82 & 48491 $\pm$ 7889 & 5.61    \\
\midrule 
\midrule

\multirow{2}{*}{\STAB{\rotatebox[origin=c]{90}{Ours}}}
& \resgnn{}      & \textbf{0.51 $\pm$ 0.01} & \textbf{-6.39} & \textbf{1.33 $\pm$ 0.08} & \textbf{-8.35} & \textbf{7.07 $\pm$ 0.20} & \textbf{-2.04} & \textbf{0.1095 $\pm$ 0.01} & \textbf{-3.42} & 46747 $\pm$ 4639 & 1.81  \\ 
& \bgnn{}        & \textbf{0.5 $\pm$ 0.01} & \textbf{-8.15} & \textbf{1.26 $\pm$ 0.08} & \textbf{-13.67} & \textbf{6.95 $\pm$ 0.21} & \textbf{-3.8} & \textbf{0.109 $\pm$ 0.01} & \textbf{-3.9} & 49222 $\pm$ 3743 & 7.2   \\ \bottomrule
\end{tabular}
}
\end{center}
\end{table*}

Our results demonstrate significant improvement of \bgnn{} over the baselines. In particular, in the heterogeneous case, \bgnn{} achieves 8\%, 14\%, 4\%, and 4\% reduction of the error for \house{}, \county{}, \vkdata{}, and \avazu{} datasets, respectively. \resgnn{} model that uses a pretrained \gbdt{} model for the input of \gnn{} also decreases RMSE, although not as much as the end-to-end model \bgnn{}. In the homogeneous dataset \wiki{}, \gbdt{} and, subsequently, \resgnn{} and \bgnn{} are outperformed by the \gnn{} model. Intuitively, when the features are homogeneous, neural network approaches are sufficient to attain the best results. This shows that \bgnn{}  \textit{leads to better qualitative results and its end-to-end training outperforms other approaches in node prediction tasks for graphs with heterogeneous tabular data}. 

We can also observe that the end-to-end combination \mlp-\gnn{} often leads to better performance than pure \textbf{GNN}. However, its improvement is smaller than for \bgnn{} which uses the advantages of GBDT models. Moreover, \gbdt{} and \lgbm{} can be effective on their own, but their performance is not stable across all datasets. Overall, these experiments demonstrate the superiority of \bgnn{} against other strong models.

\subsection{Node classification}

For node classification, we use five datasets with different properties. Due to the lack of publicly available datasets with heterogeneous node features, we adopt the datasets \textbf{House\_class} and \textbf{VK\_class} from the regression task by converting the target labels into several discrete classes. We additionally include two sparse node classification datasets \textbf{SLAP} and \textbf{DBLP} coming from heterogeneous information networks (HIN) with nodes having different types. We also include one homogeneous dataset \arxiv{} \citep{hu2020ogb}. In this dataset, the node features correspond to a 128-dimensional feature vector obtained by averaging the embeddings of words in the title and abstract. Hence, the features are not heterogeneous, and therefore GBDT is not expected to outperform neural network approaches. More details about these datasets can be found in Appendix~\ref{app:classification}.

\begin{table*}[h]
\caption{Summary of our results for node classification. Gap \% is the relative difference w.r.t.~\gat{} accuracy (the higher the better). Top-2 results are highlighted in \textbf{bold}.}\label{tab:class}
\vskip 0.15in
\begin{center}
\centering
\footnotesize
\resizebox{\textwidth}{!}{
\begin{tabular}{ll|rr|rr|rr|rr|rr}
 & & \multicolumn{8}{c|}{\textbf{\textit{Heterogeneous}}} & \multicolumn{2}{c}{\textbf{\textit{Homogeneous}}} \\ \midrule
 & &  \multicolumn{2}{c|}{\textbf{House\_class}} & \multicolumn{2}{c|}{\textbf{VK\_class}} & \multicolumn{2}{c|}{\slap} & \multicolumn{2}{c|}{\dblp} & \multicolumn{2}{c}{\arxiv} \\
 &  Method &  
 \multicolumn{1}{c}{Acc.} & \multicolumn{1}{c|}{Gap \%}  & 
 \multicolumn{1}{c}{Acc.} & \multicolumn{1}{c|}{Gap \%} &
 \multicolumn{1}{c}{Acc.} & \multicolumn{1}{c|}{Gap \%} &
 \multicolumn{1}{c}{Acc.} & \multicolumn{1}{c|}{Gap \%} &
 \multicolumn{1}{c}{Acc.} & \multicolumn{1}{c}{Gap \%} \\
 \midrule
 \midrule
\multirow{2}{*}{\STAB{\rotatebox[origin=c]{90}{\scriptsize{GBDT}}}}
& \gbdt{}        & 0.52 $\pm$ 0.01 & -16.82 & 0.57 $\pm$ 0.01 & -1.26 & 0.922 $\pm$ 0.01 & 15.12 & 0.759 $\pm$ 0.03 & -5.42 & 0.45 & -36.35   \\
& \lgbm{}        & 0.55 $\pm$ 0.00 & -11.98 & 0.579 $\pm$ 0.01 & 0.26 & \textbf{0.963 $\pm$ 0.00} & \textbf{20.3} &\textbf{ 0.913 $\pm$ 0.01} & \textbf{13.73} & 0.51 & -26.97    \\
\midrule
\multirow{4}{*}{\STAB{\rotatebox[origin=c]{90}{GNN}}}
& \gat{}          & 0.625 $\pm$ 0.00 & 0 & 0.577 $\pm$ 0.00 & 0 & 0.801 $\pm$ 0.01 & 0 & 0.802 $\pm$ 0.01 & 0                    & \textbf{0.70} & \textbf{0}  \\
& \gcn{}          & 0.6 $\pm$ 0.00 & -3.98 & 0.574 $\pm$ 0.00 & -0.6 & 0.878 $\pm$ 0.01 & 9.72 & 0.428 $\pm$ 0.04 & -46.6        & - & - \\
& \agnn{}         & 0.614 $\pm$ 0.01 & -1.73 & 0.572 $\pm$ 0.00 & -0.79 & 0.892 $\pm$ 0.01 & 11.47 & 0.794 $\pm$ 0.01 & -1.02 & - & -  \\
& \appnp{}        & 0.619 $\pm$ 0.00 & -0.89 & 0.573 $\pm$ 0.00 & -0.67 & 0.895 $\pm$ 0.01 & 11.79 & 0.83 $\pm$ 0.02 & 3.47      & - & -  \\
\midrule
\multirow{2}{*}{\STAB{\rotatebox[origin=c]{90}{NN}}}
& \mlp{}            & 0.534 $\pm$ 0.01 & -14.53 & 0.567 $\pm$ 0.01 & -1.72 & 0.759 $\pm$ 0.04 & -5.24 & 0.623 $\pm$ 0.02 & -22.3    & 0.50 & -28.91\\
& \mlp-\gnn         & \textbf{0.64 $\pm$ 0.00} & \textbf{2.36} & 0.589 $\pm$ 0.00 & 2.13 & 0.89 $\pm$ 0.01 & 11.11 & 0.81 $\pm$ 0.01 & 0.94 & \textbf{0.71} & \textbf{0.54}  \\
\midrule 
\midrule

\multirow{2}{*}{\STAB{\rotatebox[origin=c]{90}{Ours}}}
& \resgnn{}      & 0.625 $\pm$ 0.01 & -0.06 & \textbf{0.603 $\pm$ 0.00} & \textbf{4.45} & 0.905 $\pm$ 0.01 & 13.06 & \textbf{0.892 $\pm$ 0.01} & \textbf{11.11} & 0.70 & -0.33 \\ 
& \bgnn{}        & \textbf{0.682 $\pm$ 0.00} & \textbf{9.18} & \textbf{0.683 $\pm$ 0.00} & \textbf{18.3} & \textbf{0.95 $\pm$ 0.00} & \textbf{18.61} & 0.889 $\pm$ 0.01 & 10.77         & 0.67 & -4.36  \\ \bottomrule
\end{tabular}
}
\end{center}
\end{table*}

As can be seen from Table~\ref{tab:class}, on the datasets with heterogeneous tabular features (\textbf{House\_class} and \textbf{VK\_class}), \bgnn{} outperforms other approaches with a significant margin. For example, for the \textbf{VK\_class} dataset \bgnn{} achieves more than 18\% of relative increase in accuracy. This demonstrates that learned representations of GBDT together with GNN can be equally useful for node classification setting on data with heterogeneous features.

The other two datasets, \textbf{Slap} and \textbf{DBLP}, have sparse bag-of-words features that are particularly challenging for the \gnn{} model. On these two datasets, \dt{} is the strongest baseline. Moreover, since \textbf{FCNN} outperforms \gnn{}, we conclude that graph structure does not help, hence \bgnn{} is not supposed to beat \dt{}. This is indeed the case: the final accuracy of \bgnn{} is slightly worse than that of \dt{}.

In the homogeneous \arxiv{} dataset,  \mlp-\gnn{} and \gnn{} achieve the top performance followed by \resgnn{} and \bgnn{} models.\footnote{For \arxiv{} we used a different implementation of the GAT model to align results with the publicly available leaderboard: \url{https://ogb.stanford.edu/docs/nodeprop/}} In a nutshell, \textbf{GBDT} does not learn good predictions on the homogeneous input features and therefore reduces the discriminative power of \textbf{GNN}. Both cases, with sparse and with homogeneous features, show that the performance of \bgnn{} is on par or higher than of \textbf{GNN}; however, lacking heterogeneous structure in the data may make the joint training of \textbf{GBDT} and \textbf{GNN} redundant.

\subsection{Consistency across different GNN models}
\label{sec:expablation}

Seeing that our models perform significantly better than strong baselines on various datasets, we want to test whether the improvement is consistent if different GNN models are used. Thus, we ask:

\begin{enumerate}[label=Q\arabic*]
\setcounter{enumi}{2}
\item \label{item:qnns} \textit{Do different GNN models benefit from our approach of combination with GBDT?}
\end{enumerate}

To answer this question, we compare GNN models that include \gat{} \citep{velickovic2018graph}, \textbf{GCN} \citep{kipf2016semi}, \textbf{AGNN} \citep{thekumparampil2018attention}, and \textbf{APPNP} \citep{klicpera2018predict}. We substitute each of these models to \resgnn{} and \bgnn{} and measure the relative change in performance with respect to the original GNN's performance. 

\begin{figure}[h]
\centering     
\subfigure[\house]{\label{fig:a}\includegraphics[width=.32\textwidth]{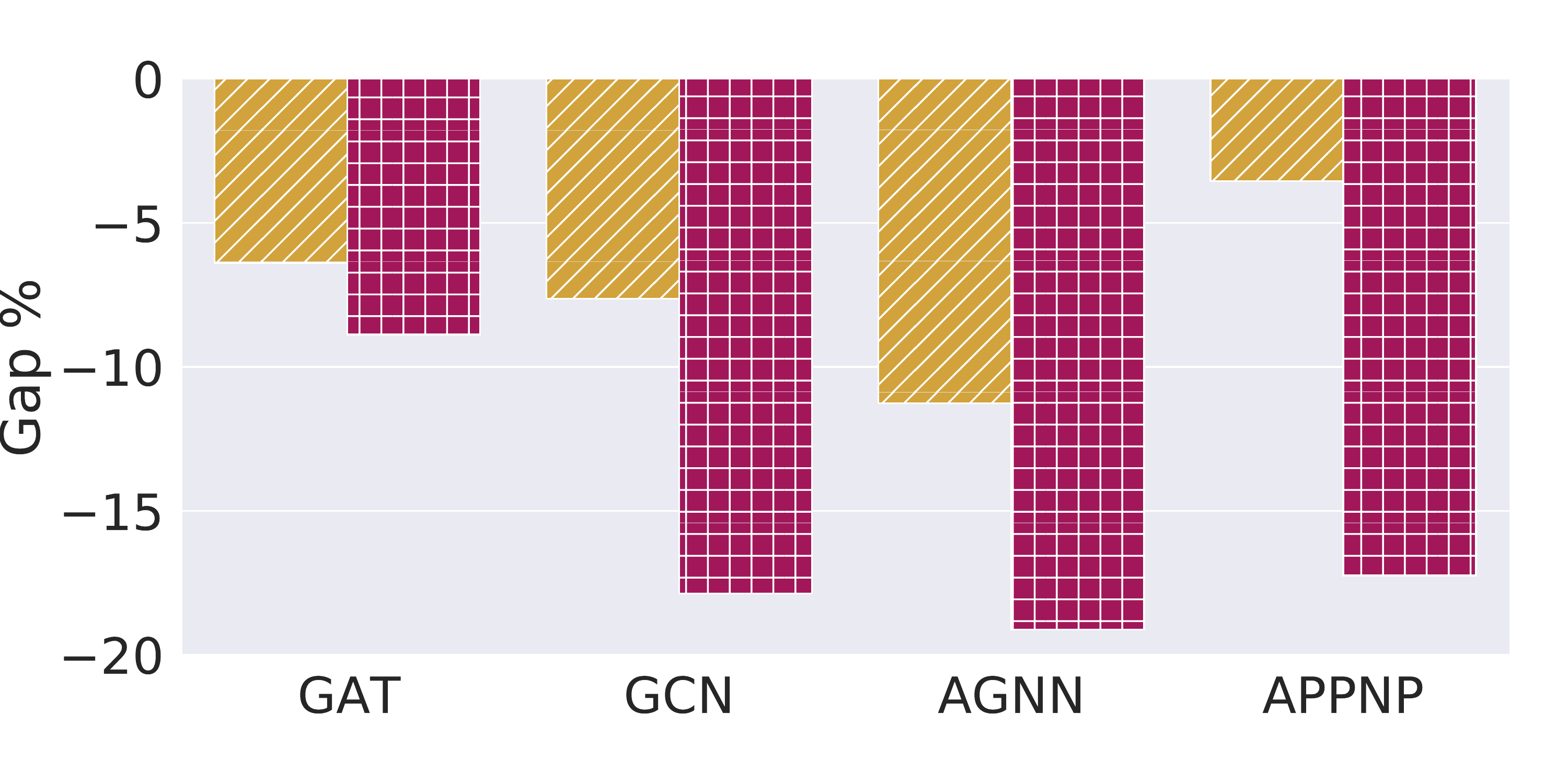}}
\subfigure[\vkdata]{\label{fig:b}\includegraphics[width=.32\textwidth]{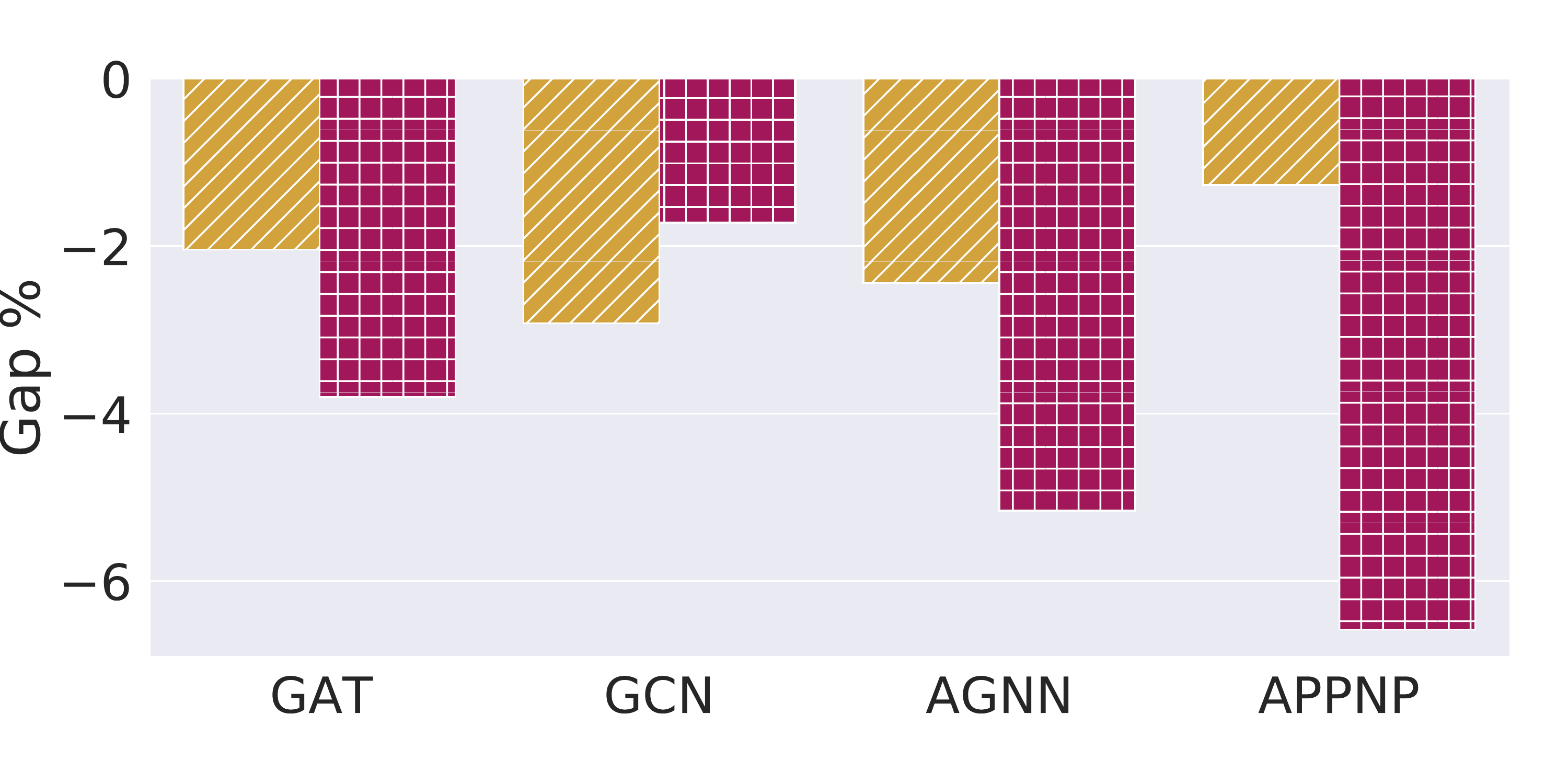}}
\subfigure[\avazu]{\label{fig:c}\includegraphics[width=.32\textwidth]{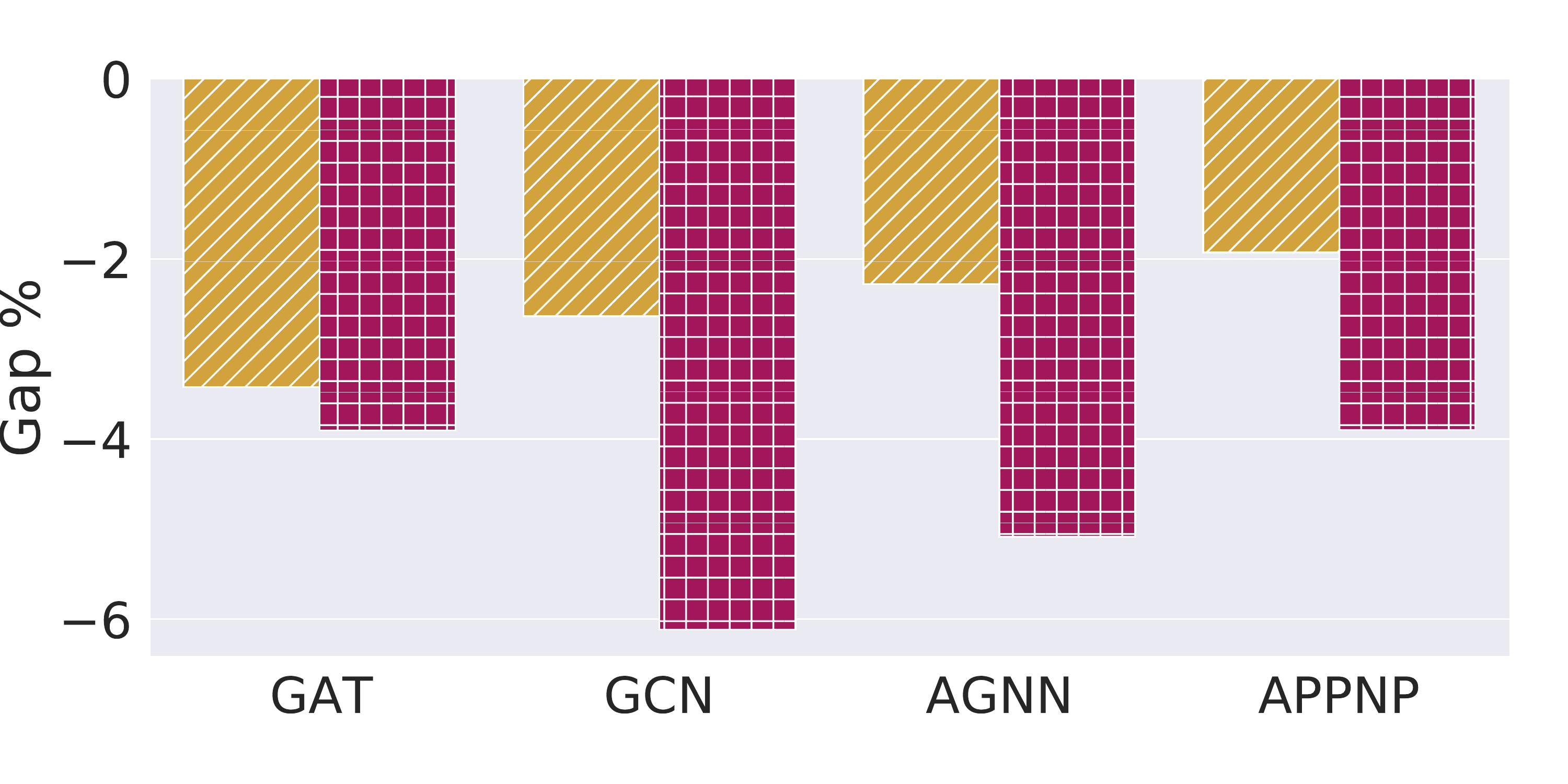}}
\caption{Relative difference for \resgnn{} (yellow, diagonal) and \bgnn{} (red, squared) for different GNN architectures w.r.t.~GNN RMSE (the smaller the better).}
\label{fig:gnns}
\end{figure}

In Figure~\ref{fig:gnns} we report the relative RMSE gap between \resgnn{} and \bgnn{} for each of the GNN architectures, i.e., we compute $gap = (r_m-r_{gnn})/r_{gnn}$, where $r_m$ and $r_{gnn}$ are RMSE of that model and of GNN respectively. This experiment positively answers \ref{item:qnns} and shows that \textit{all tested GNN architectures significantly benefit from the proposed approach}. For example, for \house{} dataset the decrease in the mean squared error is 9\%, 18\%, 19\%, and 17\% for \textbf{GAT}, \textbf{GCN}, \textbf{AGNN}, and \textbf{APPNP} models respectively. Additionally, one can see that the end-to-end training of \bgnn{} (red, squared) leads to larger improvements than a na\"ive combination of \gbdt{} and \gnn{} in \resgnn{} (yellow, diagonal). Exact metrics and training time are in Appendix~\ref{app:gnns}.

\subsection{Training time}

As the previous experiments demonstrated superior quality across various datasets and GNN models, it is important to understand if the additional GBDT part can become a bottleneck in terms of efficiency for training this model on real-world datasets. Hence, we ask:

\begin{enumerate}[label=Q\arabic*]
\setcounter{enumi}{3}
\item \label{item:runtime} \textit{Do \bgnn{} and \resgnn{} models incur a significant increase in training time?}
\end{enumerate}

To answer this question, we measure the clock time to train each model until convergence, considering early stopping. Table~\ref{tab:time} presents training time for each model. We can see that both \bgnn{} and \resgnn{} run faster than \gnn{} in most cases. In other words,  \bgnn{} and \resgnn{} models \emph{do not incur an increase in training time} but actually are more efficient than \gnn{}. For example, for \vkdata{} dataset \bgnn{} and \resgnn{} run 3x and 2x faster than \gnn{}, respectively. Moreover, \bgnn{} is consistently faster than another end-to-end implementation \mlp-\gnn{} that uses \mlp{} instead of \gbdt{} to preprocess the original input features. 






\begin{table*}[h]
\caption{Training time (s) in node regression task.}\label{tab:time}
\vskip 0.15in
\begin{center}
\centering
\footnotesize

\begin{tabular}{ll|rrrrr}
 & Method & \house & \county & \vkdata & \wiki & \avazu \\
 \midrule
 \midrule
\multirow{2}{*}{\STAB{\rotatebox[origin=c]{90}{\scriptsize{GBDT}}}}
& \gbdt  & 4 $\pm$ 1 & 2 $\pm$ 1 & 24 $\pm$ 4 & 10 $\pm$ 1 & 2 $\pm$ 2  \\
& \lgbm  & 3 $\pm$ 0 & 1 $\pm$ 0 & 5 $\pm$ 3 & 3 $\pm$ 2 & 0 $\pm$ 0 \\
\midrule
\multirow{4}{*}{\STAB{\rotatebox[origin=c]{90}{GNN}}}
& \textbf{GAT}   & 35 $\pm$ 2 & 19 $\pm$ 6 & 42 $\pm$ 4 & 15 $\pm$ 1 & 9 $\pm$ 2 \\
& \textbf{GCN}   & 28 $\pm$ 0 & 18 $\pm$ 7 & 38 $\pm$ 0 & 13 $\pm$ 3 & 12 $\pm$ 6 \\
& \textbf{AGNN}  & 38 $\pm$ 5 & 28 $\pm$ 3 & 48 $\pm$ 3 & 19 $\pm$ 5 & 14 $\pm$ 8 \\
& \textbf{APPNP} & 68 $\pm$ 1 & 34 $\pm$ 10 & 81 $\pm$ 3 & 49 $\pm$ 26 & 24 $\pm$ 15 \\
\midrule 
\multirow{2}{*}{\STAB{\rotatebox[origin=c]{90}{NN}}}
& \mlp{}         & 16 $\pm$ 5 & 2 $\pm$ 1 & 109 $\pm$ 35 & 12 $\pm$ 2 & 2 $\pm$ 0 \\
& \mlp-\gnn      & 39 $\pm$ 1 & 21 $\pm$ 6 & 48 $\pm$ 2 & 16 $\pm$ 1 & 14 $\pm$ 3 \\
\midrule 
\midrule
\multirow{2}{*}{\STAB{\rotatebox[origin=c]{90}{Ours}}}
& \resgnn{}      & 36 $\pm$ 7 & 7 $\pm$ 3 & 41 $\pm$ 7 & 31 $\pm$ 9 & 7 $\pm$ 2 \\ 
& \bgnn{}        & 20 $\pm$ 4 & 2 $\pm$ 0 & 16 $\pm$ 0 & 21 $\pm$ 7 & 5 $\pm$ 1 \\ \bottomrule
\end{tabular}
\end{center}
\end{table*}

\begin{figure}[H]
\centering     
\subfigure[House]{\label{fig:a2}\includegraphics[width=.48\textwidth]{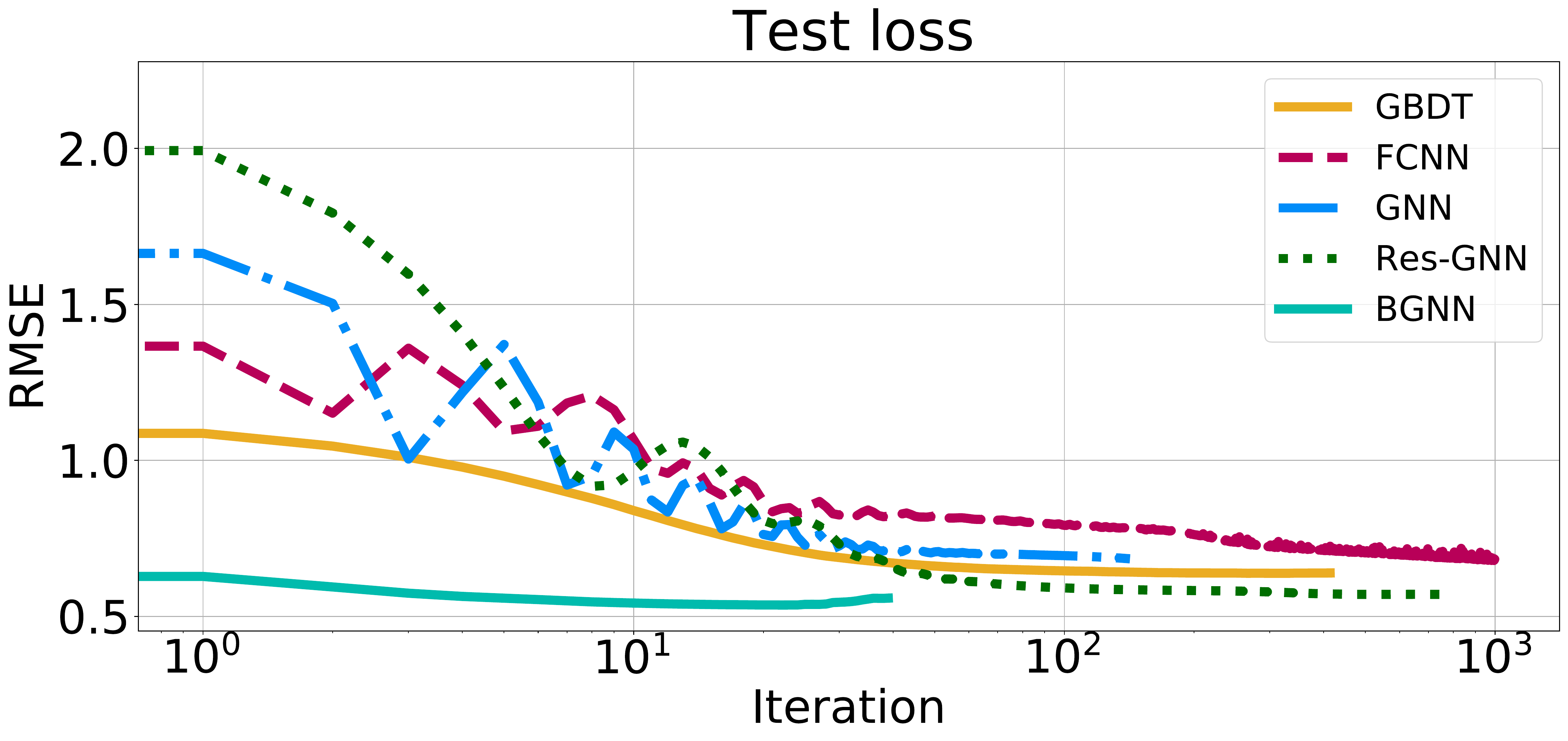}}
\subfigure[VK]{\label{fig:b2}\includegraphics[width=.48\textwidth]{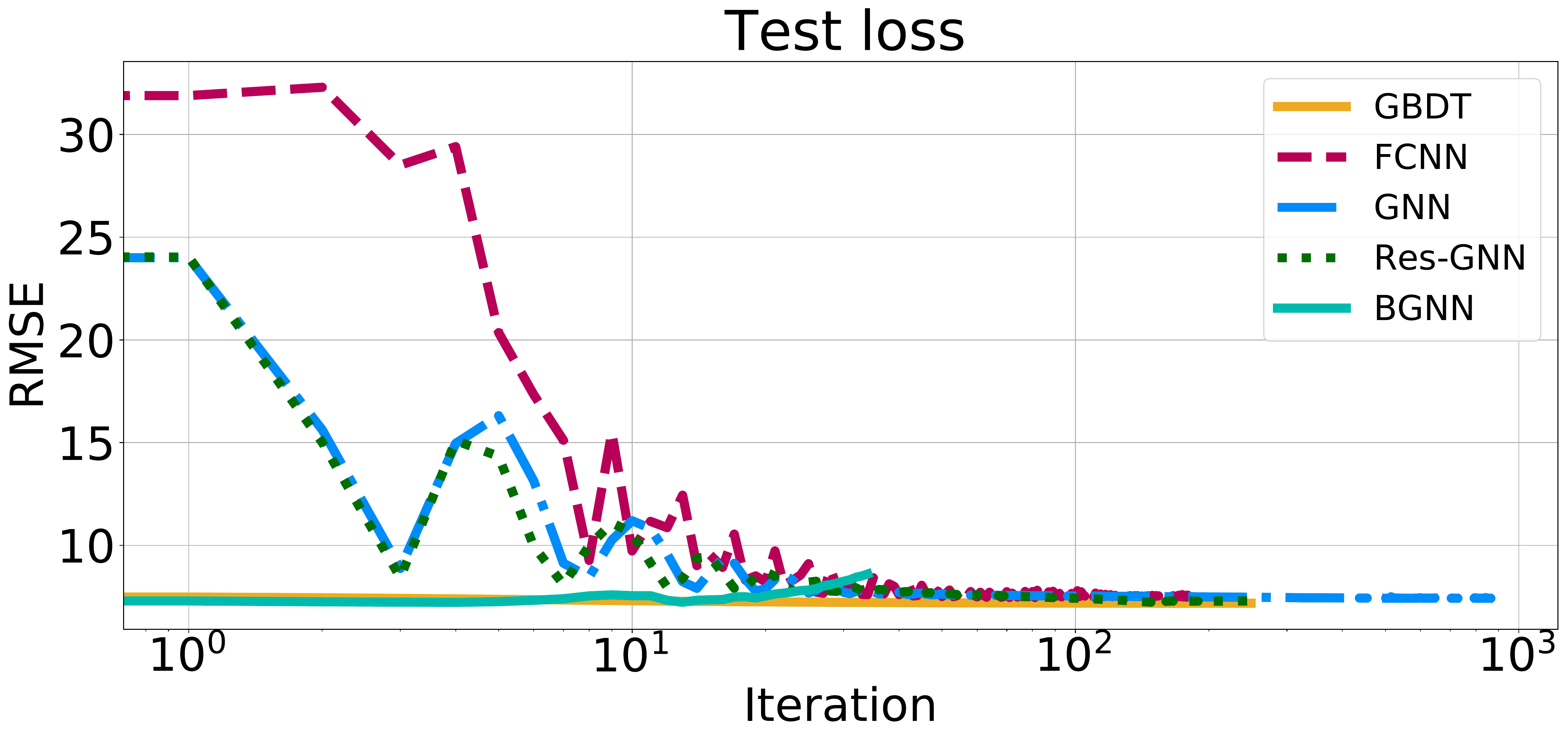}}
\caption{RMSE on the test set during training for two node regression datasets.}
\label{fig:convergence}
\end{figure}

The reason for improved efficiency is that \bgnn{} and \resgnn{} converge with a much fewer number of iterations as demonstrated in Figure~\ref{fig:convergence}. We plot RMSE on the test set during training for all models (with winning hyperparameters). We can see that \bgnn{} converges within the first ten iterations (for $k=20$), leading to fast training. In contrast, \resgnn{} is similar in terms of convergence to \gnn{} for the first 100 epochs, but then it continues decreasing RMSE unlike \gnn{} that requires much more epochs to converge. This behavior is similar for other datasets (see Appendix~\ref{app:convergence}).

\subsection{Visualizing predictions}

\begin{figure}[h]
\centering     
\subfigure[\textbf{True}]{\label{fig:intuitiona}\includegraphics[width=.3\textwidth]{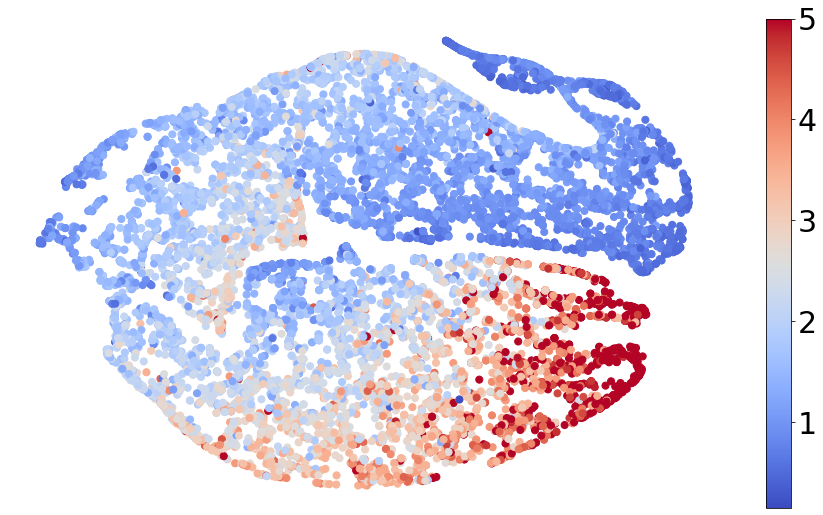}}
\subfigure[\textbf{GBDT}]{\label{fig:intuitionb}\includegraphics[width=.3\textwidth]{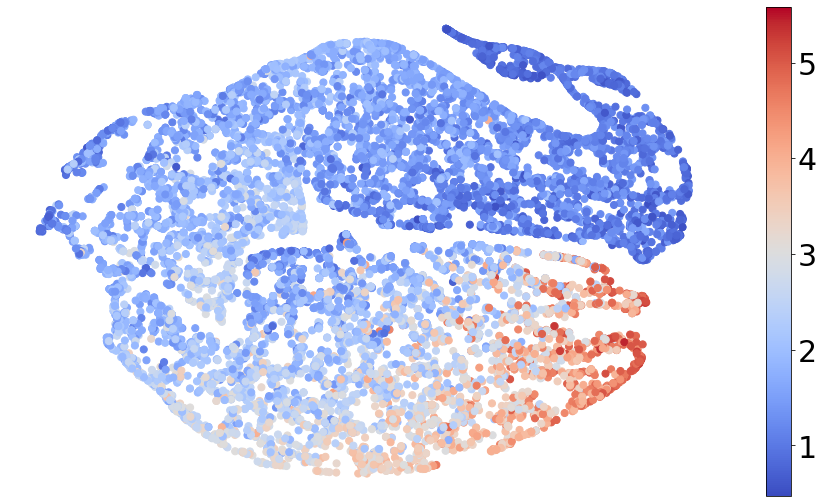}}
\subfigure[\gnn{}]{\label{fig:intuitionc}\includegraphics[width=.3\textwidth]{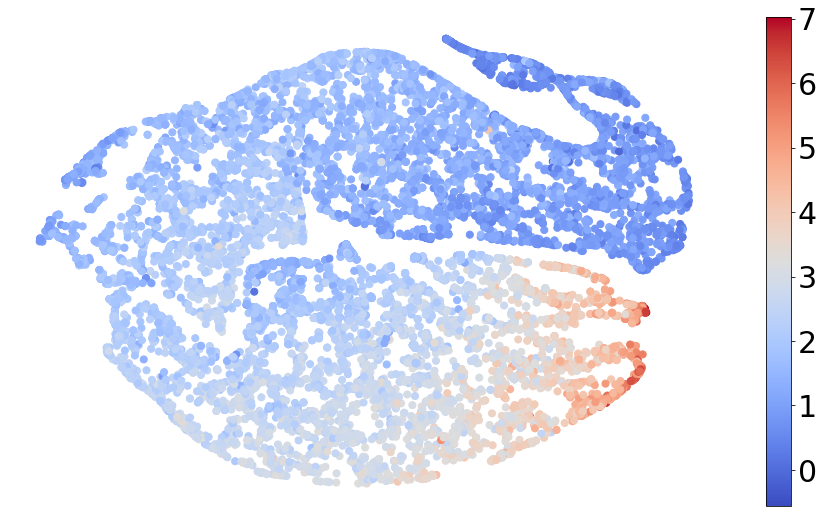}}
\subfigure[\resgnn{}]{\label{fig:intuitiond}\includegraphics[width=.3\textwidth]{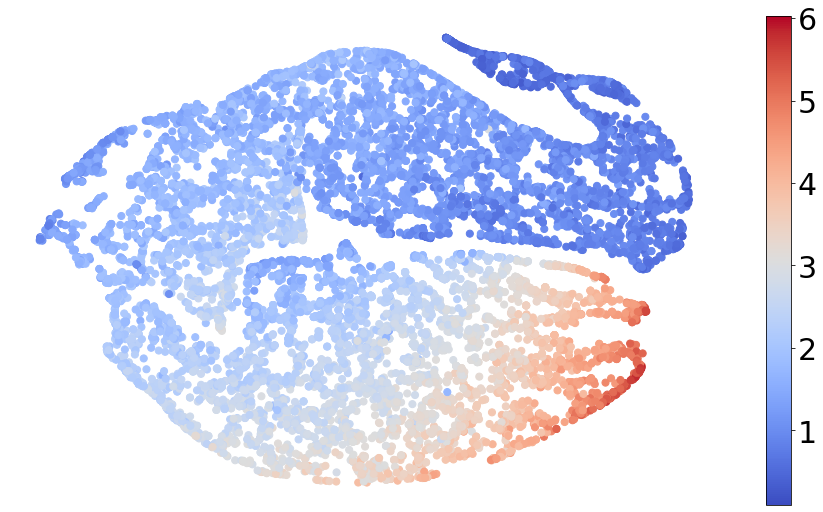}}
\subfigure[\textbf{GBDT} in \textbf{BGNN}]{\label{fig:intuitione}\includegraphics[width=.3\textwidth]{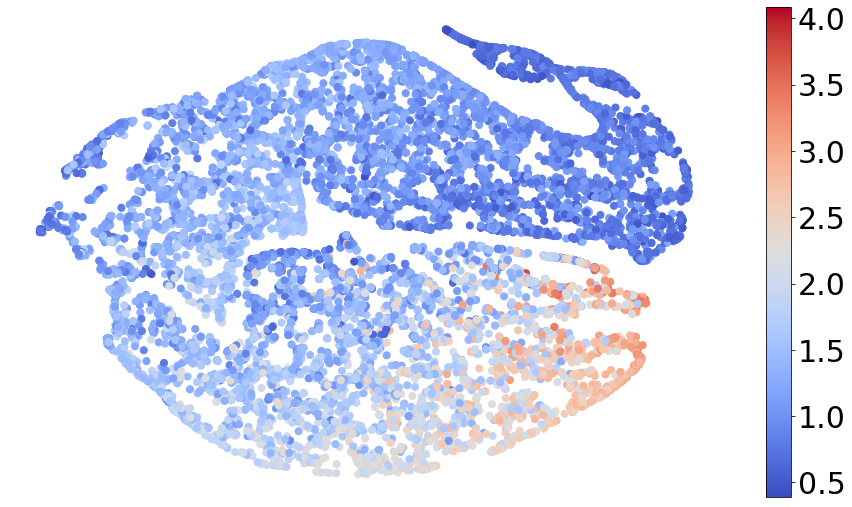}}
\subfigure[\bgnn{}]{\label{fig:intuitionf}\includegraphics[width=.3\textwidth]{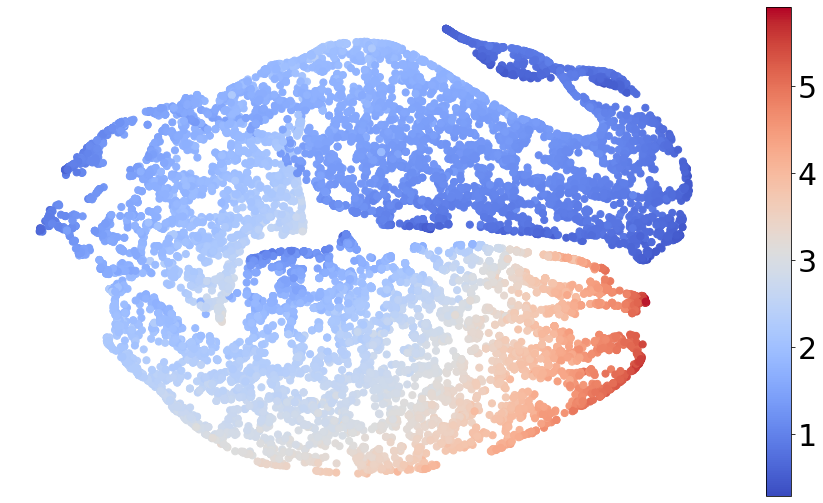}}
\caption{\house{} dataset. True labels and predictions by trained \dt{}, \gnn{}, \resgnn{}, and \bgnn{} models (training points only). Point coordinates correspond to \bgnn{} learned representations in the first hidden layer. Color represents the final predictions made by each model.}
\label{fig:intuition_house}
\end{figure}

To investigate the performance of \bgnn{}, we plot the final predictions of trained models for observations in the training set. Our motivation is to scrutinize which points are correctly classified by different models. Figure~\ref{fig:intuition_house} displays the predictions of \dt{}, \gnn{}, \resgnn{}, and \bgnn{} models as well as the true target value. To better understand the predictions of the \bgnn{} model, in Figure~\ref{fig:intuitione} we show the values predicted by \dt{} that was trained as a part of \bgnn{}. This experiment is performed on \house{} dataset, the plots for other datasets show similar trends and can be found in the supplementary materials.

Several observations can be drawn from these figures. First, the true target values change quite smoothly within local neighborhoods; however, there are a few outliers: single red points among many blue points and conversely. These points can mislead the model during the training, predicting the wrong target value for many observations in the outliers' local neighborhoods. Hence, it is important for a model to make smoothed predictions in the local neighborhoods. 

Second, comparing the prediction spaces of \dt{}, \gnn{}, \resgnn{}, and \bgnn{} models we can observe that predictions for \dt{} are much more grainy with large variations in the neighborhoods of the vertices (high quality images can be found in the supplementary materials). Intuitively, because the \dt{} model does not have access to the graph structure, it cannot propagate its predictions in the nodes' vicinity. Alternatively, \gnn{}, \resgnn{}, and \bgnn{} can extrapolate the outputs among local neighbors, smoothing out the final predictions as seen in Figures \ref{fig:intuitionc}, \ref{fig:intuitiond}, \ref{fig:intuitionf}. 

Third, focusing on the values of predictions (color bars on the right of each plot) of \dt{}, \gnn{}, and \bgnn{} models we notice that the scale of final predictions for \dt{} and \bgnn{} models is closely aligned with the true predictions, while \gnn's predictions mismatch the true values by large margin. Our intuition is that the expressive power of \dt{} to learn piecewise decision boundaries common in tabular datasets helps \dt{} and \bgnn{} to properly tune its final predictions with respect to the true range of values. In contrast, \gnn{} relies solely on neural layers to learn complex decision rules. 

Another observation comes from looking at the values predicted by \dt{} trained as a part of \bgnn{} (see Figure~\ref{fig:intuitione}). While this \dt{} model is initialized using the true target labels, it was not forced to predict the target during the training. Interestingly, this model shows the same trend and clearly captures the regions on high/low target values. On the other hand, \dt{} trained as a part of \bgnn{} is much more conservative: on all datasets, the range of predicted values is significantly smaller than the true one. We hypothesize that \dt{} is trained to scale its predictions to make them more suitable for further improvements by \gnn{}. 

\section{Conclusion}

We have presented \bgnn, a novel architecture for learning on graphs with heterogeneous tabular node features. \bgnn{} takes advantages of the GBDT model to build hyperplane decision boundaries that are common for heterogeneous data, and then utilizes GNN to refine the predictions using relational information. Our approach is end-to-end and can be incorporated with any message-passing neural network and gradient boosting method. Extensive experiments demonstrate that the proposed architecture is superior to strong existing competitors in terms of accuracy of predictions and training time. A possible direction for future research is to analyze whether this approach is profitable for graph-level predictions such as graph classification or subgraph detection.

\subsubsection*{Acknowledgments}

The authors thank the Anonymous Reviewers for their reviews and Anton Tsitsulin for kindly sharing VK data.
Liudmila Prokhorenkova also acknowledge the financial support from the Ministry of Education and Science of the Russian Federation in the framework of MegaGrant 075-15-2019-1926 and from the Russian President grant supporting leading scientific schools of the Russian Federation NSh-2540.2020.1.

\bibliography{iclr2021_conference}

\begin{thebibliography}{42}
\providecommand{\natexlab}[1]{#1}
\providecommand{\url}[1]{\texttt{#1}}
\expandafter\ifx\csname urlstyle\endcsname\relax
  \providecommand{\doi}[1]{doi: #1}\else
  \providecommand{\doi}{doi: \begingroup \urlstyle{rm}\Url}\fi

\bibitem[Arik \& Pfister(2020)Arik and Pfister]{arik2020tabnet}
Sercan~O Arik and Tomas Pfister.
\newblock Tabnet: Attentive interpretable tabular learning.
\newblock \emph{arXiv preprint arXiv:1908.07442}, 2020.

\bibitem[Badirli et~al.(2020)Badirli, Liu, Xing, Bhowmik, and
  Keerthi]{badirli2020gradient}
Sarkhan Badirli, Xuanqing Liu, Zhengming Xing, Avradeep Bhowmik, and Sathiya~S
  Keerthi.
\newblock Gradient boosting neural networks: Grownet.
\newblock \emph{arXiv preprint arXiv:2002.07971}, 2020.

\bibitem[Battaglia et~al.(2018)Battaglia, Hamrick, Bapst, Sanchez-Gonzalez,
  Zambaldi, Malinowski, Tacchetti, Raposo, Santoro, Faulkner,
  et~al.]{battaglia2018relational}
Peter~W Battaglia, Jessica~B Hamrick, Victor Bapst, Alvaro Sanchez-Gonzalez,
  Vinicius Zambaldi, Mateusz Malinowski, Andrea Tacchetti, David Raposo, Adam
  Santoro, Ryan Faulkner, et~al.
\newblock Relational inductive biases, deep learning, and graph networks.
\newblock \emph{arXiv preprint arXiv:1806.01261}, 2018.

\bibitem[Bent{\'e}jac et~al.(2020)Bent{\'e}jac, Cs{\"o}rg{\H{o}}, and
  Mart{\'\i}nez-Mu{\~n}oz]{bentejac2020comparative}
Candice Bent{\'e}jac, Anna Cs{\"o}rg{\H{o}}, and Gonzalo
  Mart{\'\i}nez-Mu{\~n}oz.
\newblock A comparative analysis of gradient boosting algorithms.
\newblock \emph{Artificial Intelligence Review}, pp.\  1--31, 2020.

\bibitem[Casas et~al.(2019)Casas, Gulino, Liao, and
  Urtasun]{casas2019spatially}
Sergio Casas, Cole Gulino, Renjie Liao, and Raquel Urtasun.
\newblock Spatially-aware graph neural networks for relational behavior
  forecasting from sensor data.
\newblock \emph{arXiv preprint arXiv:1910.08233}, 2019.

\bibitem[Clevert et~al.(2016)Clevert, Unterthiner, and Hochreiter]{elu}
Djork-Arne Clevert, Thomas Unterthiner, and Sepp Hochreiter.
\newblock Fast and accurate deep network learning by exponential linear units
  (elus).
\newblock In \emph{4th International Conference on Learning Representations},
  2016.

\bibitem[Feng et~al.(2018)Feng, Yu, and Zhou]{feng2018multi}
Ji~Feng, Yang Yu, and Zhi-Hua Zhou.
\newblock Multi-layered gradient boosting decision trees.
\newblock In \emph{Advances in neural information processing systems}, pp.\
  3551--3561, 2018.

\bibitem[Fey \& Lenssen(2019)Fey and Lenssen]{fey2019fast}
Matthias Fey and Jan~Eric Lenssen.
\newblock Fast graph representation learning with pytorch geometric.
\newblock \emph{arXiv preprint arXiv:1903.02428}, 2019.

\bibitem[Fey et~al.(2019)Fey, Lenssen, Morris, Masci, and Kriege]{fey2020deep}
Matthias Fey, Jan~E Lenssen, Christopher Morris, Jonathan Masci, and Nils~M
  Kriege.
\newblock Deep graph matching consensus.
\newblock In \emph{International Conference on Learning Representations}, 2019.

\bibitem[Friedman(2001)]{friedman2001greedy}
Jerome~H Friedman.
\newblock Greedy function approximation: a gradient boosting machine.
\newblock \emph{Annals of statistics}, pp.\  1189--1232, 2001.

\bibitem[Hazimeh et~al.(2020)Hazimeh, Ponomareva, Mol, Tan, and
  Mazumder]{hazimeh2020tree}
Hussein Hazimeh, Natalia Ponomareva, Petros Mol, Zhenyu Tan, and Rahul
  Mazumder.
\newblock The tree ensemble layer: Differentiability meets conditional
  computation.
\newblock In \emph{37th International Conference on Machine Learning (ICML
  2020)}, 2020.

\bibitem[Hu et~al.(2020{\natexlab{a}})Hu, Fey, Zitnik, Dong, Ren, Liu, Catasta,
  and Leskovec]{hu2020ogb}
Weihua Hu, Matthias Fey, Marinka Zitnik, Yuxiao Dong, Hongyu Ren, Bowen Liu,
  Michele Catasta, and Jure Leskovec.
\newblock Open graph benchmark: Datasets for machine learning on graphs.
\newblock In \emph{Conference on Neural Information Processing Systems},
  2020{\natexlab{a}}.

\bibitem[Hu et~al.(2020{\natexlab{b}})Hu, Liu, Gomes, Zitnik, Liang, Pande, and
  Leskovec]{hu2020strategies}
Weihua Hu, Bowen Liu, Joseph Gomes, Marinka Zitnik, Percy Liang, Vijay Pande,
  and Jure Leskovec.
\newblock Strategies for pre-training graph neural networks.
\newblock In \emph{International Conference on Learning Representations},
  2020{\natexlab{b}}.

\bibitem[Jia \& Benson(2020)Jia and Benson]{jia2020outcome}
Junteng Jia and Austin Benson.
\newblock Outcome correlation in graph neural network regression.
\newblock \emph{arXiv preprint arXiv:2002.08274}, 2020.

\bibitem[Kaur et~al.(2020)Kaur, Nori, Jenkins, Caruana, Wallach, and
  Wortman~Vaughan]{kaur2020interpreting}
Harmanpreet Kaur, Harsha Nori, Samuel Jenkins, Rich Caruana, Hanna Wallach, and
  Jennifer Wortman~Vaughan.
\newblock Interpreting interpretability: Understanding data scientists' use of
  interpretability tools for machine learning.
\newblock In \emph{Proceedings of the 2020 CHI Conference on Human Factors in
  Computing Systems}, 2020.

\bibitem[Ke et~al.(2017)Ke, Meng, Finley, Wang, Chen, Ma, Ye, and
  Liu]{lightgbm}
Guolin Ke, Qi~Meng, Thomas Finley, Taifeng Wang, Wei Chen, Weidong Ma, Qiwei
  Ye, and Tie-Yan Liu.
\newblock Lightgbm: A highly efficient gradient boosting decision tree.
\newblock In \emph{Advances in Neural Information Processing Systems 30}. 2017.

\bibitem[Keriven \& Peyr{\'e}(2019)Keriven and Peyr{\'e}]{keriven2019universal}
Nicolas Keriven and Gabriel Peyr{\'e}.
\newblock Universal invariant and equivariant graph neural networks.
\newblock In \emph{Advances in Neural Information Processing Systems}, pp.\
  7092--7101, 2019.

\bibitem[Kipf \& Welling(2017)Kipf and Welling]{kipf2016semi}
Thomas~N Kipf and Max Welling.
\newblock Semi-supervised classification with graph convolutional networks.
\newblock In \emph{International Conference on Learning Representations}, 2017.

\bibitem[Klicpera et~al.(2019)Klicpera, Bojchevski, and
  G{\"u}nnemann]{klicpera2018predict}
Johannes Klicpera, Aleksandar Bojchevski, and Stephan G{\"u}nnemann.
\newblock Predict then propagate: Graph neural networks meet personalized
  pagerank.
\newblock In \emph{International Conference on Learning Representations}, 2019.

\bibitem[Li et~al.(2019)Li, Qin, Wang, and Metzler]{li2019combining}
Pan Li, Zhen Qin, Xuanhui Wang, and Donald Metzler.
\newblock Combining decision trees and neural networks for learning-to-rank in
  personal search.
\newblock In \emph{Proceedings of the 25th ACM SIGKDD International Conference
  on Knowledge Discovery \& Data Mining}, pp.\  2032--2040, 2019.

\bibitem[Loukas(2020)]{loukas2020what}
Andreas Loukas.
\newblock What graph neural networks cannot learn: depth vs width.
\newblock In \emph{International Conference on Learning Representations}, 2020.

\bibitem[Maron et~al.(2019)Maron, Fetaya, Segol, and
  Lipman]{maron2019universality}
Haggai Maron, Ethan Fetaya, Nimrod Segol, and Yaron Lipman.
\newblock On the universality of invariant networks.
\newblock In \emph{International conference on machine learning}, pp.\
  4363--4371. PMLR, 2019.

\bibitem[Mazyavkina et~al.(2020)Mazyavkina, Sviridov, Ivanov, and
  Burnaev]{mazyavkina2020reinforcement}
Nina Mazyavkina, Sergey Sviridov, Sergei Ivanov, and Evgeny Burnaev.
\newblock Reinforcement learning for combinatorial optimization: A survey.
\newblock \emph{arXiv preprint arXiv:2003.03600}, 2020.

\bibitem[Pace \& Barry(1997)Pace and Barry]{pace1997sparse}
R~Kelley Pace and Ronald Barry.
\newblock Sparse spatial autoregressions.
\newblock \emph{Statistics \& Probability Letters}, 33\penalty0 (3):\penalty0
  291--297, 1997.

\bibitem[Peters et~al.(2019)Peters, Niculae, and Martins]{peters2019sparse}
Ben Peters, Vlad Niculae, and Andr{\'e}~FT Martins.
\newblock Sparse sequence-to-sequence models.
\newblock In \emph{Proceedings of the 57th Annual Meeting of the Association
  for Computational Linguistics}, pp.\  1504--1519, 2019.

\bibitem[Popov et~al.(2019)Popov, Morozov, and Babenko]{popov2019neural}
Sergei Popov, Stanislav Morozov, and Artem Babenko.
\newblock Neural oblivious decision ensembles for deep learning on tabular
  data.
\newblock In \emph{International Conference on Learning Representations}, 2019.

\bibitem[Prokhorenkova et~al.(2018)Prokhorenkova, Gusev, Vorobev, Dorogush, and
  Gulin]{prokhorenkova2018catboost}
Liudmila Prokhorenkova, Gleb Gusev, Aleksandr Vorobev, Anna~Veronika Dorogush,
  and Andrey Gulin.
\newblock Catboost: unbiased boosting with categorical features.
\newblock In \emph{Advances in neural information processing systems}, pp.\
  6638--6648, 2018.

\bibitem[Ren et~al.(2019)Ren, Liu, Huang, Dai, Bo, and
  Zhang]{ren2019heterogeneous}
Yuxiang Ren, Bo~Liu, Chao Huang, Peng Dai, Liefeng Bo, and Jiawei Zhang.
\newblock Heterogeneous deep graph infomax.
\newblock \emph{arXiv preprint arXiv:1911.08538}, 2019.

\bibitem[Rozemberczki et~al.(2019)Rozemberczki, Allen, and
  Sarkar]{rozemberczki2019multiscale}
Benedek Rozemberczki, Carl Allen, and Rik Sarkar.
\newblock Multi-scale attributed node embedding.
\newblock \emph{arXiv preprint arXiv:1909.13021}, 2019.

\bibitem[Satorras \& Estrach(2018)Satorras and Estrach]{garcia2018fewshot}
Victor~Garcia Satorras and Joan~Bruna Estrach.
\newblock Few-shot learning with graph neural networks.
\newblock In \emph{International Conference on Learning Representations}, 2018.

\bibitem[Song et~al.(2019)Song, Shi, Xiao, Duan, Xu, Zhang, and
  Tang]{song2019autoint}
Weiping Song, Chence Shi, Zhiping Xiao, Zhijian Duan, Yewen Xu, Ming Zhang, and
  Jian Tang.
\newblock Autoint: Automatic feature interaction learning via self-attentive
  neural networks.
\newblock In \emph{Proceedings of the 28th ACM International Conference on
  Information and Knowledge Management}, pp.\  1161--1170, 2019.

\bibitem[Stokes et~al.(2020)Stokes, Yang, Swanson, Jin, Cubillos-Ruiz, Donghia,
  MacNair, French, Carfrae, Bloom-Ackerman, et~al.]{stokes2020deep}
Jonathan~M Stokes, Kevin Yang, Kyle Swanson, Wengong Jin, Andres Cubillos-Ruiz,
  Nina~M Donghia, Craig~R MacNair, Shawn French, Lindsey~A Carfrae, Zohar
  Bloom-Ackerman, et~al.
\newblock A deep learning approach to antibiotic discovery.
\newblock \emph{Cell}, 180\penalty0 (4):\penalty0 688--702, 2020.

\bibitem[Sun et~al.(2020)Sun, Guo, Zhang, Zhang, Regol, Hu, Guo, Tang, Yuan,
  He, and Coates]{sun20framework}
Jianing Sun, Wei Guo, Dengcheng Zhang, Yingxue Zhang, Florence Regol, Yaochen
  Hu, Huifeng Guo, Ruiming Tang, Han Yuan, Xiuqiang He, and Mark Coates.
\newblock A framework for recommending accurate and diverse items using
  bayesian graph convolutional neural networks.
\newblock In \emph{Proceedings of the 26th ACM SIGKDD International Conference
  on Knowledge Discovery \& Data Mining}, pp.\  2030--2039, 2020.

\bibitem[Sun et~al.(2019)Sun, Lin, and Zhu]{sun2019adagcn}
Ke~Sun, Zhouchen Lin, and Zhanxing Zhu.
\newblock Adagcn: Adaboosting graph convolutional networks into deep models.
\newblock \emph{arXiv preprint arXiv:1908.05081}, 2019.

\bibitem[Thekumparampil et~al.(2018)Thekumparampil, Wang, Oh, and
  Li]{thekumparampil2018attention}
Kiran~K Thekumparampil, Chong Wang, Sewoong Oh, and Li-Jia Li.
\newblock Attention-based graph neural network for semi-supervised learning.
\newblock \emph{arXiv preprint arXiv:1803.03735}, 2018.

\bibitem[Tsitsulin et~al.(2018)Tsitsulin, Mottin, Karras, and
  M{\"u}ller]{tsitsulin2018verse}
Anton Tsitsulin, Davide Mottin, Panagiotis Karras, and Emmanuel M{\"u}ller.
\newblock Verse: Versatile graph embeddings from similarity measures.
\newblock In \emph{Proceedings of the 2018 World Wide Web Conference}, pp.\
  539--548, 2018.

\bibitem[Veli{\v{c}}kovi{\'{c}} et~al.(2018)Veli{\v{c}}kovi{\'{c}}, Cucurull,
  Casanova, Romero, Li{\`{o}}, and Bengio]{velickovic2018graph}
Petar Veli{\v{c}}kovi{\'{c}}, Guillem Cucurull, Arantxa Casanova, Adriana
  Romero, Pietro Li{\`{o}}, and Yoshua Bengio.
\newblock {Graph Attention Networks}.
\newblock \emph{International Conference on Learning Representations}, 2018.

\bibitem[Wang et~al.(2020)Wang, Jamnik, and Lio]{wang2020abstract}
Duo Wang, Mateja Jamnik, and Pietro Lio.
\newblock Abstract diagrammatic reasoning with multiplex graph networks.
\newblock In \emph{International Conference on Learning Representations}, 2020.

\bibitem[Wu et~al.(2020)Wu, Pan, Zhou, Chang, and Zhu]{wu20unsupervised}
Man Wu, Shirui Pan, Chuan Zhou, Xiaojun Chang, and Xingquan Zhu.
\newblock Unsupervised domain adaptive graph convolutional networks.
\newblock In \emph{Proceedings of The Web Conference}, 2020.

\bibitem[Xiao et~al.(2019)Xiao, Zhang, Yang, and Zhai]{xiao2019non}
Yuxin Xiao, Zecheng Zhang, Carl Yang, and Chengxiang Zhai.
\newblock Non-local attention learning on large heterogeneous information
  networks.
\newblock In \emph{2019 IEEE International Conference on Big Data (Big Data)},
  pp.\  978--987. IEEE, 2019.

\bibitem[Yang et~al.(2018)Yang, Morillo, and Hospedales]{yang2018deep}
Yongxin Yang, Irene~Garcia Morillo, and Timothy~M Hospedales.
\newblock Deep neural decision trees.
\newblock \emph{arXiv preprint arXiv:1806.06988}, 2018.

\bibitem[Zhou \& Feng(2019)Zhou and Feng]{zhou2019deep}
Zhi-Hua Zhou and Ji~Feng.
\newblock Deep forest.
\newblock \emph{National Science Review}, 6\penalty0 (1):\penalty0 74--86,
  2019.

\end{thebibliography}
\bibliographystyle{iclr2021_conference}
\newpage
\appendix
\section{Further related work}
\label{app:related}

To the best of our knowledge, there are no approaches combining the benefits of GBDT and GNN models for representation learning on graphs with tabular data. However, there are many attempts to adapt non-graph neural networks for tabular data or to combine them with gradient boosting in different ways.

Several works \citep{popov2019neural, yang2018deep, zhou2019deep, feng2018multi, hazimeh2020tree} attempt to mitigate the non-differentiable nature of decision trees. For example, \cite{popov2019neural} proposed to replace hard choices for tree splitting features and splitting thresholds with their continuous counterparts, using $\alpha$-entmax transformation~\citep{peters2019sparse}. While such an approach becomes suitable for a union of decision trees with GNN, the computational burden of training both end-to-end becomes a bottleneck for large graphs. 

Another method \citep{badirli2020gradient} uses neural networks as weak learners for the GBDT model. For graph representation problems such as node regression, one can replace standard neural networks with graph neural networks. However, training different GNN as weak classifiers at once would be exhaustive. Additionally, such a combination lacks some advantages of GBDT, like handling heterogeneous and categorical features and missing values. An approach called AdaGCN~\citep{sun2019adagcn} incorporates AdaBoost ideas into the design of GNNs in order to construct deep models. Again, this method does not exploit the advantages of GBDT methods.

Finally, \citet{li2019combining} investigated different ways of combining decision-tree-based models and neural networks. While the motivation is similar to ours~--- get the benefits of both types of models~--- the paper focuses specifically on learning-to-rank problems. Additionally, while some of their methods are similar in spirit to \resgnn{}, they do not update GBDT in an end-to-end manner, which is a substantial contribution of the current research.

\section{Hyperparameters}
\label{app:hyper}
Parameters in brackets \{\} are selected by hyperparameter search on the validation set. 

\lgbm{}: number of leaves is \{15, 63\}, $||\lambda||_2=0$, boosting type is gbdt, number of epochs is 1000, early stopping rounds is 100. 

\gbdt{}: depth is \{4, 6\}, $||\lambda||_2=0$, number of epochs is 1000, early stopping rounds is 100.  

\mlp{}: number of layers is \{2, 3\}, dropout is \{0., 0.5\}, hidden dimension is 64, number of epochs is 5000, early stopping rounds is 2000. 

\gnn{}: dropout rate is \{0., 0.5\}, hidden dimension is 64,  number of epochs is 2000, early stopping rounds is 200. \gat{}, \gcn{}, and \agnn{} models have two convolutional layers with dropout and ELU activation function \citep{elu}. \appnp{} has a two-layer fully-connected neural network with dropout and ELU activation followed by a convolutional layer with $k=10$ and $\alpha=0.1$. We use eight heads with eight hidden neurons for \gat{} model. 

\resgnn{}: dropout rate is \{0., 0.5\}, hidden dimension is 64, number of epochs is 1000, early stopping rounds is 100. We also tune whether to use solely predictions of \gbdt{} model or append them to the input features. \gbdt{} model is trained for 1000 epochs. 

\bgnn{}: dropout rate is \{0., 0.5\}, hidden dimension is 64,  number of epochs is 200, early stopping rounds is 10, number of trees and backward passes per epoch is \{10, 20\}, depth of the tree is 6. We also tune whether to use solely predictions of \gbdt{} model or append them to the input features.

For all models, we also perform a hyperparameter search on learning rate in \{0.1, 0.01\}. Every hyperparameter setting is evaluated three times and an average is taken. We use five random splits for train/validation/test with 0.6/0.2/0.2 ratio. The average across five seeds is reported in the tables. 

\section{Regression datasets}
\label{app:datasets}

In \textbf{House} dataset \citep{pace1997sparse}, nodes are the properties, edges connect the proximal nodes, and the target is the property's price. We use the publicly available dataset~\citep{pace1997sparse} of all the block groups in California collected from the 1990 Census. We connect each block with at most five of its nearest neighbors if they lie within a ball of a certain radius, as measured by latitude and longitude. We keep the following node features: \textit{MedInc, HouseAge, AveRooms, AveBedrms, Population, AveOccup}. 

\textbf{County} dataset \citep{jia2020outcome} is a county-level election map network. Each node is a county, and two nodes are connected if they share a border. We consider node features coming from the 2016 year. These features include \textit{DEM, GOP, MedianIncome, MigraRate, BirthRate, DeathRate, BachelorRate, UnemploymentRate}. We follow the setup of the original paper and select UnemploymentRate as the target label. We filter out all nodes in the original data if they do not have features. 

\textbf{VK} dataset \citep{tsitsulin2018verse} comes from a popular social network where people are mutually connected based on friendships, and the regression problem is to predict the age of a person. We use an open-access subsample of the VK social network of the first 1M users.\footnote{\url{https://github.com/xgfs/vk-userinfo }} Then, the dataset has been preprocessed to keep only the users who opt in to share their demographic information and preferences: \textit{country, city, has\_mobile, last\_seen\_platform, political, religion\_id, alcohol, smoking, relation, sex, university}. 

\textbf{Wiki} dataset \citep{rozemberczki2019multiscale} represents a page-page network on a specific topic (squirrels) with the task of predicting average monthly traffic. The features are bag-of-words for informative nouns (3148 in total) that appeared in the main text of the Wikipedia article. The target is the average monthly traffic between October 2017 and November 2018 for each article. 

\textbf{Avazu} dataset \citep{song2019autoint} represents a device-device network, with two devices being connected if they appear on the same site within the same application. For this dataset, the goal is to predict click-through-rate (CTR) for each device. We take the first 10M rows from the publicly available train log of user clicks.\footnote{\url{https://www.kaggle.com/c/avazu-ctr-prediction}} We compute CTR for each device id and filter those ids that do not have at least 10 ad displays. We connect two devices if they had ad displays on the same site id from the same application id. The node features are anonymized categories: \textit{C1, C14, C15, C16, C17, C18, C19, C20, C21}.

\section{Classification datasets}
\label{app:classification}

For node classification, we consider three types of node features: heterogeneous (\vkdata{} and \house{}), sparse (\slap{} and \dblp), and homogeneous (\arxiv).

For \house{} and \vkdata{}, we transform the original numerical target value with respect to the bin it falls to. More specifically, for \vkdata{} we consider the classes $<20, 20-25, 25-30, \ldots, 45-50, >50$ for the age attribute. Similarly, for \house{} dataset we replace the target value with the bin it falls to in the range $[1, 1.5, 2, 2.5]$. Hence, there are 7 and 5 classes for \vkdata{} and \house{}, respectively.

\begin{wraptable}[11]{r}{6.5cm}
\vspace{-5pt}
\caption{Summary of classification datasets.}\label{tab:dataclass}
\vspace{-5pt}
\begin{center}
\centering
\footnotesize
\begin{tabular}{lrrr}
\toprule
 & \textbf{SLAP} & \textbf{DBLP} & \arxiv{}  \\
 \midrule
 \midrule
\textbf{\# Nodes}         &  20419  &	14475 & 169343	 \\
\textbf{\# Edges}         &  172248 &	40269 & 1166243 \\ 
\textbf{\# Features} &  2701   &	5002  & 128 \\ 
\textbf{Classes}   & 15	& 4 & 40 \\
\textbf{Min Class} & 103	& 745 & 29 \\
\textbf{Max Class} & 534	& 1197 & 27321 \\
\bottomrule
\end{tabular}
\end{center}
\end{wraptable}

For datasets with sparse features, we consider two datasets coming from heterogeneous information networks (HIN), where nodes have a few different types. A common way to represent HIN is through meta-paths, i.e., a collection of all possible paths between nodes of a particular type. For example, for a citation network, one may specify paths of the type paper-author-paper (PAP) and the type paper-subject-paper (PSP). Then the original graph is approximated as several adjacency matrices for different types. 

\textbf{DBLP} dataset \citep{ren2019heterogeneous} is a network with three node types (authors, papers, conferences) and four target classes of the authors (database, data mining, information retrieval, and machine learning). 
To obtain a single graph, we use the adjacency matrix for the relation APA, which closely reflects the relationships between authors. 
Each author has a bag-of-words representation (300 words) of all the abstracts published by the author. Furthermore, for every node, we compute the degrees for all types of relations and append them as additional node features. Namely, we have two additional node features corresponding to degrees for paper nodes in APA and APCPA adjacency matrices. 

\textbf{SLAP} dataset \citep{xiao2019non} is a multiple-hub network in bioinformatics that contains node types such as chemical compound, gene, disease, pathway, etc. The goal is to predict one of 15 gene types. To obtain a single graph, we use the adjacency matrix for the relation GG between genes. Each gene has 3000 features that correspond to the extracted gene ontology terms (GO terms). As for \dblp, we compute the degrees for all types of relations and append them as additional node features. 

As a dataset with homogeneous node features we consider \arxiv{} \citep{hu2020ogb}. The node features correspond to a 128-dimensional feature vector obtained by averaging the embeddings of words in the title and abstract. 
Note that for this particular dataset we used the implementation of GAT\footnote{\url{https://github.com/Espylapiza/dgl/blob/master/examples/pytorch/ogb/ogbn-arxiv/models.py}} as a backbone architecture for \gnn{}, \resgnn{}, and \bgnn{} models. This model scored the top place on the leaderboard.\footnote{\url{https://ogb.stanford.edu/docs/leader_nodeprop/\#ogbn-arxiv}}
A summary of statistics for all datasets is outlined in Table~\ref{tab:dataclass}.

\section{Comparison of GNN models }
\label{app:gnns}

In this section, we show the exact RMSE values and time for all tested GNN models on all regression datasets. We consider several state-of-the-art GNN models that include \gat{} \citep{velickovic2018graph}, \textbf{GCN} \citep{kipf2016semi}, \textbf{AGNN} \citep{thekumparampil2018attention}, and \textbf{APPNP} \citep{klicpera2018predict}. 

Table~\ref{tab:appgnns} demonstrates that for all considered models \bgnn{} and \resgnn{} achieve significant increase in performance compared to vanilla \gnn{}. Additionally, end-to-end training of \bgnn{} achieves typically better results than a straightforward implementation of \resgnn{}.

\begin{table*}[h]
\caption{Summary of our results for different GNN architectures for node regression.} 
\label{tab:appgnns}
\vskip 0.15in
\begin{center}
\centering
\footnotesize
\resizebox{\textwidth}{!}{
\begin{tabular}{ll|rr|rr|rr|rr|rr}
 & & \multicolumn{2}{c|}{\textbf{House}} & \multicolumn{2}{c|}{\textbf{County}} & \multicolumn{2}{c|}{\textbf{VK}} & \multicolumn{2}{c|}{\wiki} & \multicolumn{2}{c}{\avazu}\\
 & Method &  \multicolumn{1}{c}{RMSE} & \multicolumn{1}{c|}{Time (s)}  & 
 \multicolumn{1}{c}{RMSE} & \multicolumn{1}{c|}{Time (s)} &
 \multicolumn{1}{c}{RMSE} & \multicolumn{1}{c|}{Time (s)} &
 \multicolumn{1}{c}{RMSE} & \multicolumn{1}{c|}{Time (s)} &
 \multicolumn{1}{c}{RMSE} & \multicolumn{1}{c}{Time (s)} \\
 \midrule
 \midrule
 \multirow{3}{*}{\STAB{\rotatebox[origin=c]{90}{\gat}}}
& \gnn{}        &    0.54 $\pm$ 0.01 & 35 $\pm$ 2&	1.45 $\pm$ 0.06 & 19 $\pm$ 6  &	7.22 $\pm$ 0.19 & 42 $\pm$ 4 &	45916 $\pm$ 4527 & 15 $\pm$ 1 &	0.113 $\pm$ 0.01 & 9 $\pm$ 2 \\
& \resgnn{}     &    0.51 $\pm$ 0.01 & 36 $\pm$ 7&	1.33 $\pm$ 0.08 & 7 $\pm$ 3	  &  7.07 $\pm$ 0.20 & 41 $\pm$ 7 &	46747 $\pm$ 4639 & 31 $\pm$ 9 &	0.109 $\pm$ 0.01 & 7 $\pm$ 2 \\
& \bgnn{}       &    0.5 $\pm$ 0.01 & 20 $\pm$ 4&	1.26 $\pm$ 0.08 & 2 $\pm$ 0	  &  6.95 $\pm$ 0.21 & 16 $\pm$ 0 &	49222 $\pm$ 3743 & 21 $\pm$ 7 &	0.109 $\pm$ 0.01 & 5 $\pm$ 1\\ \midrule
\multirow{3}{*}{\STAB{\rotatebox[origin=c]{90}{\gcn}}}
& \gnn{}        & 0.63 $\pm$ 0.01 & 28 $\pm$ 0 &	1.48 $\pm$ 0.08 & 18 $\pm$ 7 &	7.25 $\pm$ 0.19 & 38 $\pm$ 0 &	44936 $\pm$ 4083 & 13 $\pm$ 3	& 0.114 $\pm$ 0.02 & 12 $\pm$ 6  \\
& \resgnn{}     & 0.59 $\pm$ 0.01 & 25 $\pm$ 2 &	1.35 $\pm$ 0.09 & 11 $\pm$ 5 &	7.03 $\pm$ 0.20 & 52 $\pm$ 6	 &  44876 $\pm$ 3777 & 21 $\pm$ 5	& 0.111 $\pm$ 0.02 & 9 $\pm$ 6  \\
& \bgnn{}       & 0.54 $\pm$ 0.01 & 41 $\pm$ 15 &	1.33 $\pm$ 0.13 & 12 $\pm$ 8 &	7.12 $\pm$ 0.21 & 76 $\pm$ 6 &	47426 $\pm$ 4112 & 22 $\pm$ 11	& 0.107 $\pm$ 0.01 & 4 $\pm$ 1 \\ 

\midrule
\multirow{3}{*}{\STAB{\rotatebox[origin=c]{90}{\agnn}}}
& \gnn{}        &     0.59 $\pm$ 0.01 & 38 $\pm$ 5	& 1.45 $\pm$ 0.08 & 28 $\pm$ 3	& 7.26 $\pm$ 0.20 & 48 $\pm$ 3	& 45982 $\pm$ 3058 & 19 $\pm$ 5	  & 0.113 $\pm$ 0.02 & 14 $\pm$ 8  \\
& \resgnn{}     &     0.52 $\pm$ 0.01 & 33 $\pm$ 4	& 1.3 $\pm$ 0.07 & 16 $\pm$ 4	& 7.08 $\pm$ 0.20 & 51 $\pm$ 15	& 46010 $\pm$ 2355 & 24 $\pm$ 3	  & 0.111 $\pm$ 0.02 & 7 $\pm$ 2 \\
& \bgnn{}       &     0.49 $\pm$ 0.01 & 34 $\pm$ 4	& 1.28 $\pm$ 0.08 & 3 $\pm$ 1	& 6.89 $\pm$ 0.21 & 25 $\pm$ 4	& 53080 $\pm$ 5117 & 47 $\pm$ 37  & 0.108 $\pm$ 0.02 & 5 $\pm$ 1 \\ \midrule
\multirow{3}{*}{\STAB{\rotatebox[origin=c]{90}{\appnp}}}
& \gnn{}        & 0.69 $\pm$ 0.01 & 68 $\pm$ 1	& 1.5 $\pm$ 0.11 & 34 $\pm$ 10	& 13.23 $\pm$ 0.12 & 81 $\pm$ 3	    & 53426 $\pm$ 4159 & 49 $\pm$ 26 & 0.113 $\pm$ 0.01 & 24 $\pm$ 15 \\
& \resgnn{}     & 0.67 $\pm$ 0.01 & 58 $\pm$ 12	& 1.41 $\pm$ 0.12 & 19 $\pm$ 10	& 13.06 $\pm$ 0.17 & 76 $\pm$ 11	& 53206 $\pm$ 4593 & 66 $\pm$ 27 & 0.110 $\pm$ 0.01 & 15 $\pm$ 10 \\
& \bgnn{}       & 0.59 $\pm$ 0.01 & 21 $\pm$ 7	& 1.33 $\pm$ 0.10 & 17 $\pm$ 6	& 12.36 $\pm$ 0.14 & 50 $\pm$ 6	    & 54359 $\pm$ 4734 & 30 $\pm$ 13 & 0.108 $\pm$ 0.01 & 6 $\pm$ 1 \\

 \bottomrule
\end{tabular}
}
\end{center}
\end{table*}

\newpage
\section{Loss convergence}
\label{app:convergence}

\begin{figure}[!htb]
\centering     
\subfigure[\county]{\label{fig:a3}\includegraphics[width=.32\textwidth]{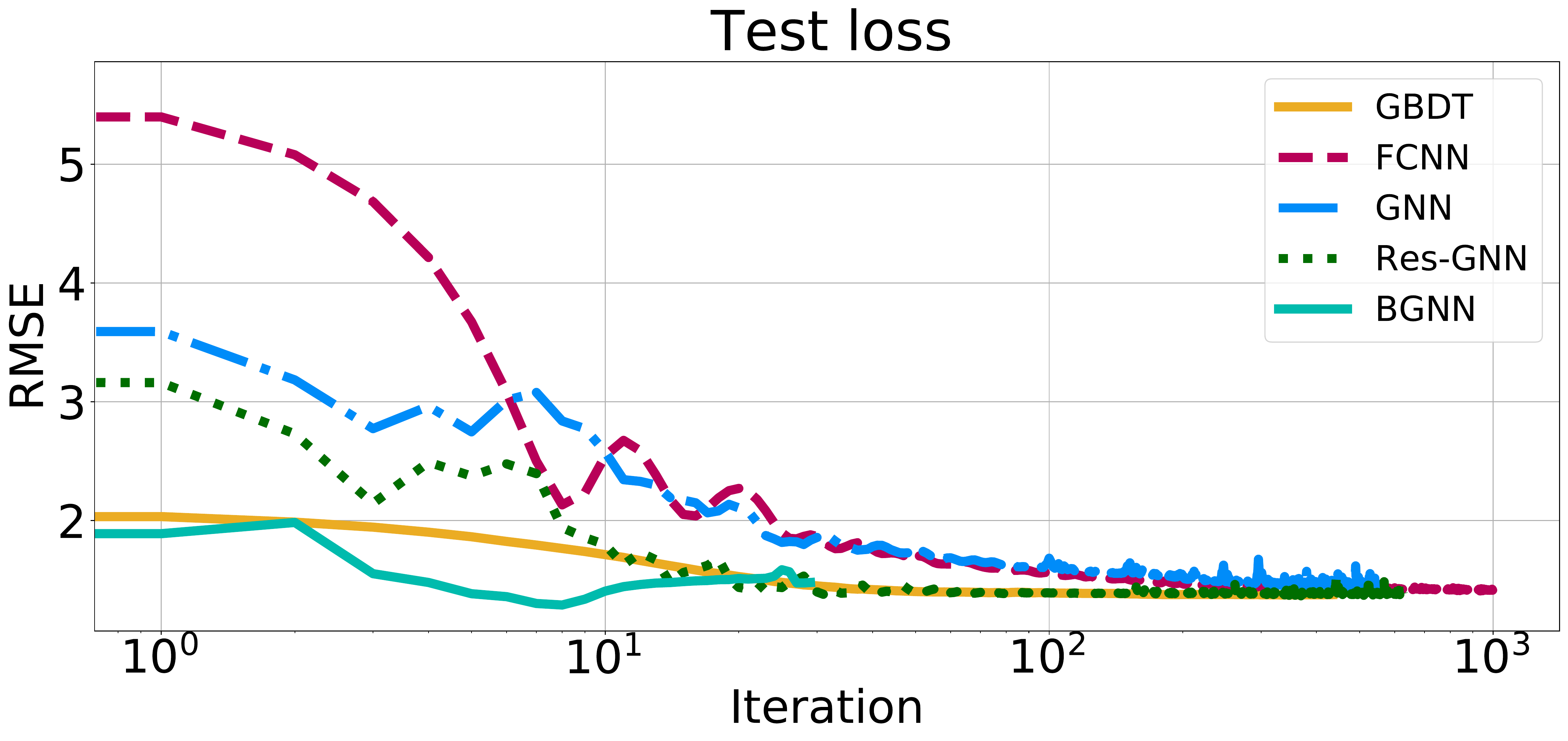}}
\subfigure[\wiki]{\label{fig:b3}\includegraphics[width=.32\textwidth]{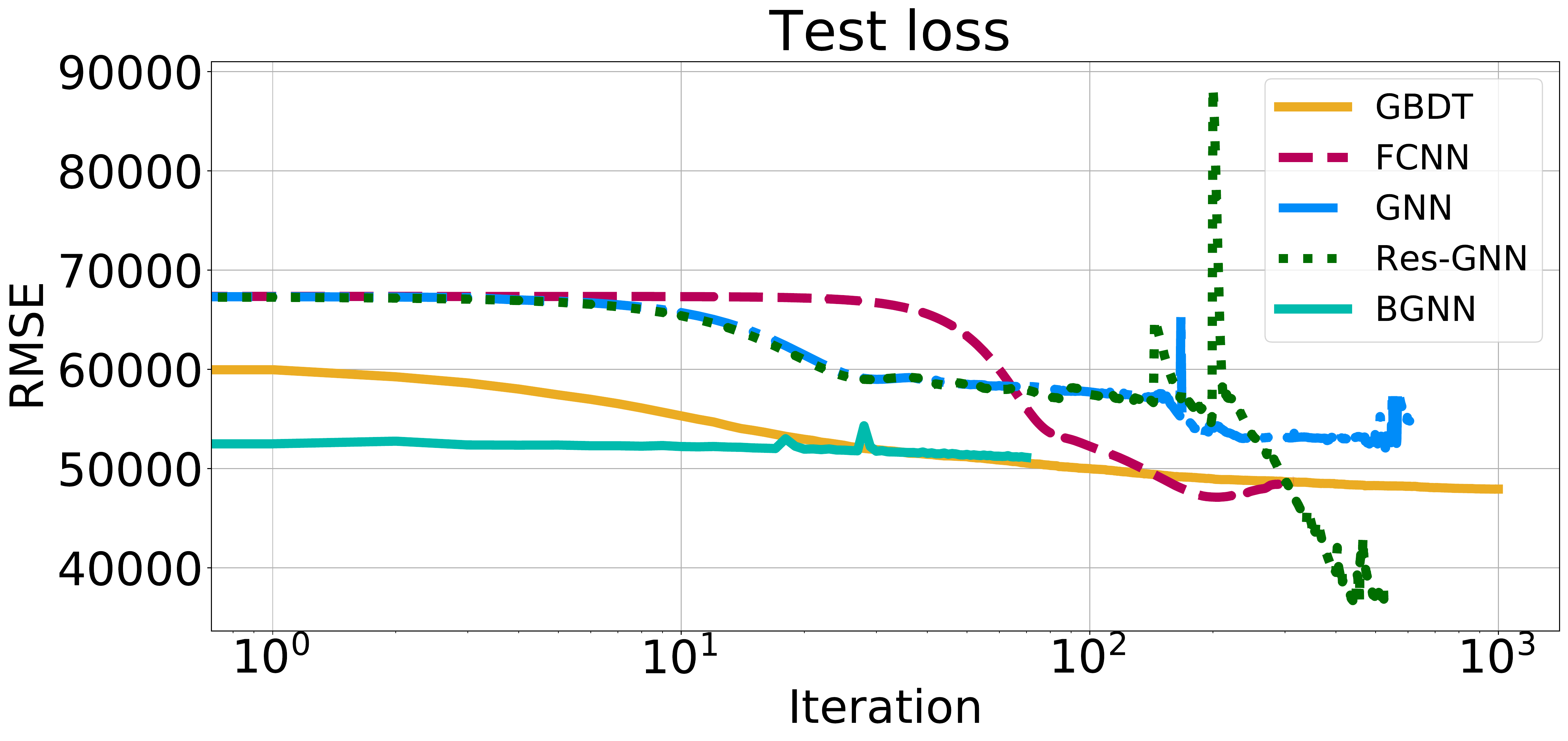}}
\subfigure[\avazu]{\label{fig:c3}\includegraphics[width=.32\textwidth]{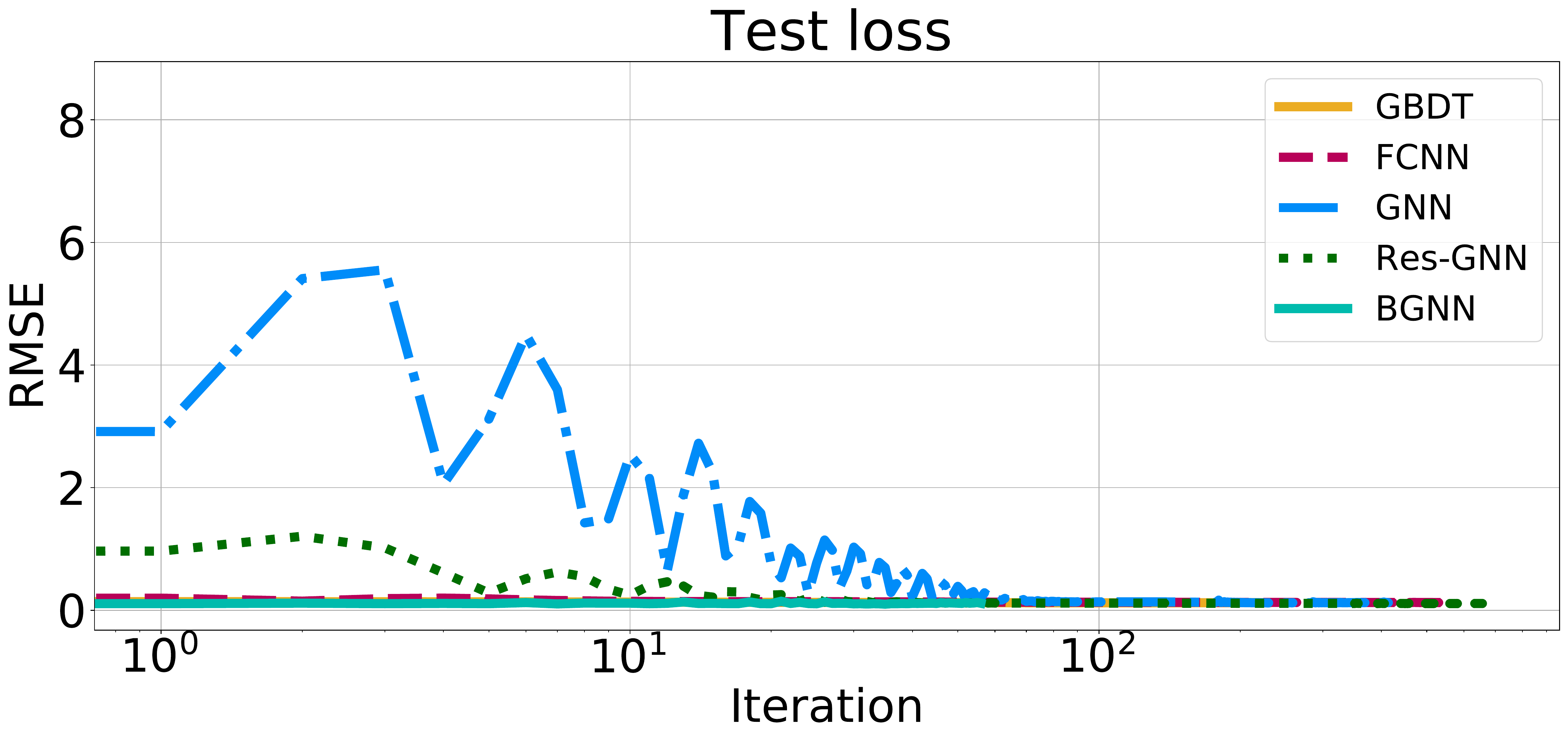}}
\caption{Summary of RMSE of test set during training for node regression datasets.}
\label{fig:convergence2}
\end{figure}

In Figure~\ref{fig:convergence2}, we plot RMSE on the test set during training for the remaining datasets~--- \county{}, \wiki{}, and \avazu{}. These results confirm that \bgnn{} converges to its optimal value within the first ten iterations (for $k=20$). Note that on the \wiki{} dataset, similarly to Figure~\ref{fig:convergence}, \resgnn{} convergence is similar to \gnn{} for the first 100 iterations and then the loss of \resgnn{} decreases faster than of \gnn{}. 

\end{document}